\documentclass[10pt,twocolumn,letterpaper]{article}

\usepackage{iccv}
\usepackage{times}
\usepackage{epsfig}
\usepackage{graphicx}
\usepackage{amsthm,amsmath,amssymb}
\usepackage{mathrsfs}
\usepackage{bm}
\usepackage{multirow}
\usepackage{paralist}
\usepackage{overpic}

\def\blu#1{\textbf{\color{blue} #1}} %in Table
\def\red#1{\textbf{\color{red}\underline{#1}}} %in Table

\usepackage[pagebackref=true,breaklinks=true,letterpaper=true,colorlinks,bookmarks=false]{hyperref}

\iccvfinalcopy % *** Uncomment this line for the final submission

\def\iccvPaperID{8501} % *** Enter the ICCV Paper ID here

\ificcvfinal\fi

\begin{document}

%%%%%%%%% TITLE
\title{Visual Saliency Transformer}

\author{
	Nian Liu$^{1}$\footnotemark[1]
	\hspace{25pt}
	Ni Zhang$^{2}$\footnotemark[1]
	\hspace{25pt}
	Kaiyuan Wan$^{2}$
	\hspace{25pt}
	Ling Shao$^{1}$
	\hspace{25pt}
	Junwei Han$^{2}$\footnotemark[2]
	\hspace{25pt}
	\\
	$^1$Inception Institute of Artificial Intelligence
	\hspace{8pt}
	$^2$Northwestern Polytechnical University
	\\
	{\tt\small
    \{liunian228, nnizhang.1995, kaiyuan.wan0106, junweihan2010\}@gmail.com, ling.shao@ieee.org
    }
}

\maketitle
\footnotetext[1]{Equal contribution.}
\footnotetext[2]{Corresponding author.}

\ificcvfinal\thispagestyle{empty}\fi

%%%%%%%%% ABSTRACT
\begin{abstract}
    Existing state-of-the-art saliency detection methods heavily rely on CNN-based architectures. 
    Alternatively, we rethink this task from a convolution-free sequence-to-sequence perspective and predict saliency by modeling long-range dependencies, which can not be achieved by convolution.
    Specifically, we develop a novel unified model based on a pure transformer, namely, Visual Saliency Transformer (VST), for both RGB and RGB-D salient object detection (SOD). It takes image patches as inputs and leverages the transformer to propagate global contexts among image patches.
    Unlike conventional architectures used in Vision Transformer (ViT), we leverage multi-level token fusion and propose a new token upsampling method under the transformer framework to get high-resolution detection results. 
    We also develop a token-based multi-task decoder to simultaneously perform saliency and boundary detection by introducing task-related tokens and a novel patch-task-attention mechanism.
    Experimental results show that our model outperforms existing methods on both RGB and RGB-D SOD benchmark datasets. Most importantly, our whole framework not only provides a new perspective for the SOD field but also shows a new paradigm for transformer-based dense prediction models. Code is available at \url{https://github.com/nnizhang/VST}.
\end{abstract}
%%%%%%%%% BODY TEXT
\section{Introduction}
SOD aims to detect objects that attract peoples' eyes and can help many vision tasks, \eg, \cite{shimoda2016distinct,gan2015devnet}.
Recently, RGB-D SOD has also gained growing interest with the extra spatial structure information from the depth data.
Current state-of-the-art SOD methods are dominated by convolutional architectures \cite{lecun1998gradient}, on both RGB and RGB-D data. They often adopt an encoder-decoder CNN architecture \cite{noh2015learning,ronneberger2015unet}, where the encoder encodes the input image to multi-level features and the decoder integrates the extracted features to predict the final saliency map. Based on this simple architecture, most efforts have been made to build a powerful decoder for predicting better saliency results. To this end, they introduced various attention models \cite{liu2018picanet,zhang2018pagr,chen2020dpanet}, multi-scale feature integration methods \cite{hou2018dss,MINet-CVPR2020,fan2020bbsnet,luo2020Cas-Gnn}, and multi-task learning frameworks \cite{wang2018salient,zhang2019capsal,zhao2019EGNet,CVPR2020_LDF,Wei2020CoNet}. An additional demand for RGB-D SOD is to effectively fuse cross-modal information, \ie, the appearance information and the depth cues. Existing works propose various modality fusion methods, such as feature fusion \cite{han2017cnns,chen2018progressively,fan2020bbsnet,Fu2020JLDCF,specificity_rgbd_sod}, knowledge distillation \cite{piao2020a2dele}, dynamic convolution \cite{HDFNet-ECCV2020}, attention models \cite{Li2020CMWNet,zhang2020ATSA}, and graph neural networks \cite{luo2020Cas-Gnn}.
Hence, CNN-based methods have achieved impressive results \cite{wang2019salient1,zhou2021rgb}.

However, all previous methods are limited in learning global long-range dependencies. Global contexts \cite{goferman2011context,zhao2015saliency,ren2015exploiting,luo2017non,liu2018picanet} and global contrast \cite{zhai2006visual,borji2012exploiting,cheng2014global} have been proved crucial for saliency detection for a long time. Nevertheless, due to the intrinsic limitation of CNNs that they extract features in local sliding windows, previous methods can hardly exploit the crucial global cues.
Although some methods utilized fully connected layers \cite{liu2016dhsnet,han2017cnns}, global pooling layers \cite{luo2017non,liu2018picanet,wang2017stagewise}, and non-local modules \cite{liu2020S2MA,chen2020dpanet} to incorporate the global context, they only did such in certain layers and the standard CNN-based architecture remains unchanged.

Recently, Transformer \cite{vaswani2017attention} was proposed to model global long-range dependencies among word sequences for machine translation. The core idea is the self-attention mechanism, which leverages the query-key correlation to relate different positions in a sequence. Transformer stacks the self-attention layers multiple times in both encoder and decoder, thus can model long-range dependencies in every layer.
Hence, it is natural to introduce the Transformer to SOD, leveraging the global cues in the model all the way.

In this paper, for the first time, we rethink SOD from a new sequence-to-sequence perspective and develop a novel unified model for both RGB and RGB-D SOD based on a pure transformer, which is named Visual Saliency Transformer.
We follow the recently proposed ViT models \cite{dosovitskiy2020image,yuan2021tokens} to divide each image into patches and adopt the Transformer model on the patch sequence. Then, the Transformer propagates long-range dependencies between image patches, without any need of using convolution.
However, it is not straightforward to apply ViT for SOD. On the one hand, how to perform dense prediction tasks based on pure transformer still remains an open question. On the other hand, ViT usually tokenizes the image to a very coarse scale. How to adapt ViT to the high-resolution prediction demand of SOD is also unclear.

To solve the first problem, we design a token-based transformer decoder by introducing task-related tokens to learn decision embeddings. Then, we propose a novel patch-task-attention mechanism to generate dense-prediction results, which provides a new paradigm for using transformer in dense prediction tasks. Motivated by previous SOD models \cite{zhao2019EGNet,Zhou2020ITSD,zhang2020select,Wei2020CoNet} that leveraged boundary detection to boost the SOD performance, we build a multi-task decoder to simultaneously conduct saliency and boundary detection by introducing a saliency token and a boundary token. This strategy simplifies the multitask prediction workflow by simply learning task-related tokens, thus largely reduces the computational costs while obtaining better results. To solve the second problem, inspired by the Tokens-to-Token (T2T) transformation \cite{yuan2021tokens}, which reduces the length of tokens, we propose a new reverse T2T transformation to upsample tokens by expanding each token into multiple sub-tokens. Then, we upsample patch tokens progressively and fuse them with low-level tokens to obtain the final full-resolution saliency map.
In addition,
we also use a cross modality transformer to deeply explore the interaction between multi-modal information for RGB-D SOD. Finally, our VST outperforms existing state-of-the-art SOD methods with a comparable number of parameters and computational costs, on both RGB and RGB-D data.

Our main contributions can be summarized as follows:
\begin{compactitem}
\item For the first time, we design a novel unified model based on the pure transformer architecture for both RGB and RGB-D SOD, from a new perspective of sequence-to-sequence modeling.

\item We design a multi-task transformer decoder to jointly conduct saliency and boundary detection by introducing task-related tokens and patch-task-attention.

\item We propose a new token upsampling method for transformer-based framework.

\item Our proposed VST model achieves state-of-the-art results on both RGB and RGB-D SOD benchmark datasets, which demonstrates its effectiveness and the potential of transformer-based models for SOD.

\end{compactitem}

%-------------------------------------------------------------------------
\section{Related Work}

\subsection{Deep Learning Based SOD}
CNN-based approaches have become a mainstream trend in both RGB and RGB-D SOD and achieved promising performance.
Most methods \cite{hou2018dss,wang2017stagewise,MINet-CVPR2020,GateNet,fan2020bbsnet} leveraged a multi-level feature fusion strategy
by using UNet \cite{ronneberger2015unet} or HED-style \cite{xie2015hed} network structures.
Some works introduced the attention mechanism to learn more discriminative features, including spatial and channel attention \cite{Piao2019dmra,zhang2018pagr,fan2020bbsnet,chen2020dpanet} or pixel-wise contextual attention \cite{liu2018picanet}.
Other works \cite{liu2016dhsnet,wang2018rfcn,deng2018r3net,liu2019salient,chen2020PGAR} tried to design recurrent networks to refine the saliency map step-by-step.
In addition, some works introduced multi-task learning, \eg, fixation prediction \cite{wang2018salient}, image caption \cite{zhang2019capsal}, and edge detection \cite{qin2019basnet, zhao2019EGNet,CVPR2020_LDF,zhang2020select,Wei2020CoNet} to boost the SOD performance.

As for RGB-D SOD, 
many methods have designed various models to fuse RGB and depth features and obtained significant results.
Some models \cite{chen2018progressively, chen2019three, Fu2020JLDCF} adopted simple feature fusion methods, \ie, concatenation, summation, or multiplication.
Some others \cite{zhao2019contrast, li2020icnet, Piao2019dmra, Li2020CMWNet} leveraged the depth cues to generate spatial or channel attention to enhance the RGB features.
Besides, dynamic convolution \cite{HDFNet-ECCV2020}, graph neural networks \cite{luo2020Cas-Gnn}, and knowledge distillation \cite{piao2020a2dele} were also adopted to implement multi-modal feature fusion.
In addition, \cite{liu2020S2MA, liu2020ReDWeb-S, chen2020dpanet} adopted the cross-attention mechanism to propagate long-range cross-modal interactions between RGB and depth cues.

Different from previous CNN-based methods, we are the first to rethink SOD from a sequence-to-sequence perspective and propose a unified model based on pure transformer 
for both RGB and RGB-D SOD. In our model, we follow \cite{qin2019basnet, zhao2019EGNet,CVPR2020_LDF,zhang2020select,Wei2020CoNet} to leverage boundary detection to boost the SOD performance. However, different from these CNN-based models, we design a novel token-based multitask decoder to achieve this goal under the transformer framework.

\subsection{Transformers in Computer Vision}
Vaswani \etal \cite{vaswani2017attention} first proposed a transformer encoder-decoder architecture for machine translation, where multi-head self-attention and point-wise feed-forward layers are stacked multiple times.
Recently, more and more works have introduced the Transformer model to various computer vision tasks and achieved excellent results. 
Some works combined CNNs and transformers into hybrid architectures for object detection \cite{carion2020end,zhu2020deformable}, panoptic segmentation \cite{wang2020maxdeeplab}, lane shape prediction \cite{liu2021end}, and so on.
Typically, they first use CNNs to extract image features and then leverage the Transformer to incorporate long-range dependencies.

\begin{figure*}[!t]
  \graphicspath{{Figures/Network/}}
  \centering
  \includegraphics[width=1\linewidth]{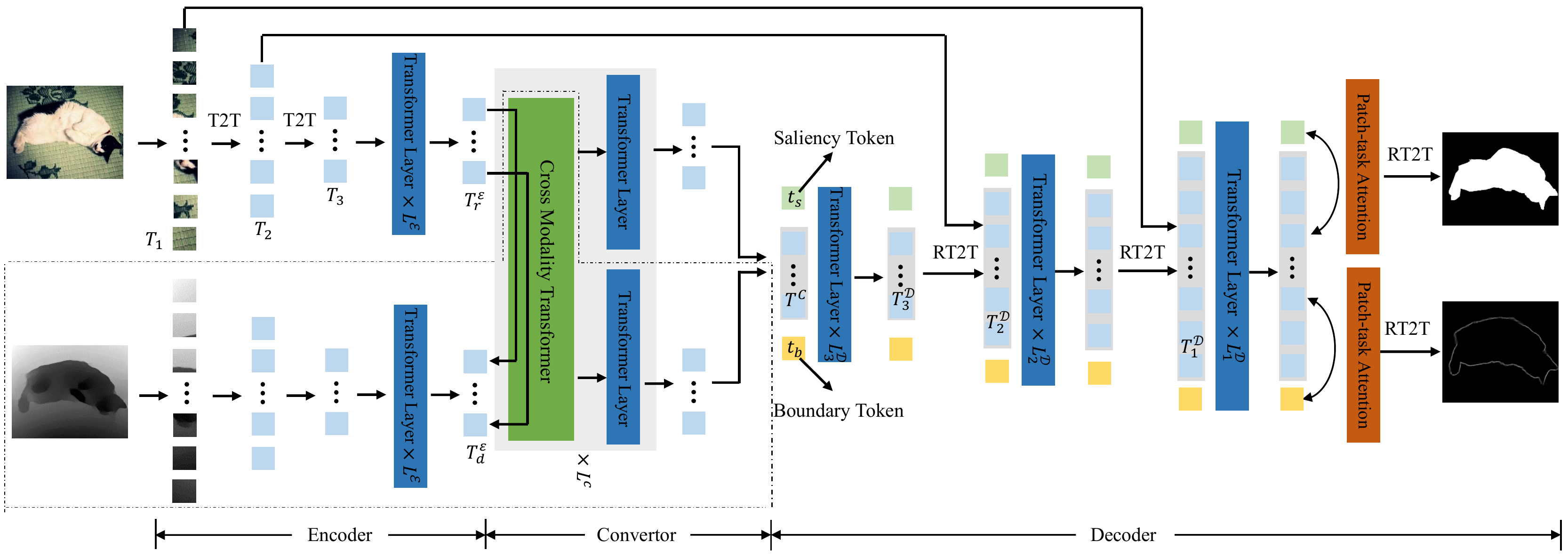}
  \caption{Overall architecture of our proposed VST model for both RGB and RGB-D SOD. It first uses an encoder to generate multi-level tokens from the input image patch sequence. Then, a convertor is adopted to convert the patch tokens to the decoder space, and also performs cross-modal information fusion for RGB-D data. Finally, a decoder simultaneously predicts the saliency map and the boundary map via the proposed task-related tokens and the patch-task-attention mechanism.
  An RT2T transformation is also proposed to progressively upsample patch tokens.
The dotted line represents exclusive components for RGB-D SOD.}
  \label{VST}
  \vspace{-0.4cm}
\end{figure*}

Other works design pure transformer models to process images from the sequence-to-sequence perspective. ViT \cite{dosovitskiy2020image} divided each image into a sequence of flattened 2D patches and then adopted the Transformer for image classification.
Touvron \etal \cite{touvron2020training} introduced a teacher-student strategy 
to improve the data-efficiency of ViT and Wang \etal \cite{wang2021pvt} proposed a pyramid architecture to adapt ViT for dense prediction tasks.
T2T-ViT \cite{yuan2021tokens} adopted the T2T module to model local structures, thus generating multiscale token features.
In this work, we adopt T2T-ViT as the backbone and propose a novel multitask decoder and a reverse T2T token upsampling method. It is noteworthy that our usage of task-related tokens is different from previous models. 
In \cite{dosovitskiy2020image,touvron2020training}, the class token is directly used for image classification via adopting a multilayer perceptron on the token embedding. 
However, 
we can not obtain dense prediction results directly from a single task token. Thus, we propose to perform patch-task-attention between patch tokens and the task tokens to predict saliency and boundary maps. We believe our strategy will also inspire future transformer models for other dense prediction tasks.

Another related work to ours is \cite{zheng2020rethinking}, which introduces transformer into the semantic segmentation task.
The authors adopted a vision transformer as a backbone and then reshaped the token sequences to 2D image features.
Then, they predicted full-resolution segmentation maps using convolution and bilinear upsampling. 
Their model still falls into the hybrid architecture category. In contrast, our model is a pure transformer architecture and does not rely on any convolution operation and bilinear upsampling.

%-------------------------------------------------------------------------
\section{Visual Saliency Transformer}
Figure~\ref{VST} shows the overall architecture of our proposed VST model.
The main components include a transformer encoder based on T2T-ViT, a transformer convertor to convert patch tokens from the encoder space to the decoder space, and a multi-task transformer decoder.

\subsection{Transformer Encoder}
Similar to other CNN-based SOD methods, which often utilize pretrained image classification models such as VGG \cite{simonyan2014vgg} and ResNet \cite{he2016resnet} as the backbone of their encoders to extract image features, we adopt the pretrained T2T-ViT \cite{yuan2021tokens} model as our backbone, as detailed below.

\begin{figure*}[!t]
  \graphicspath{{Figures/Network/}}
  \centering
  \includegraphics[width=1\linewidth]{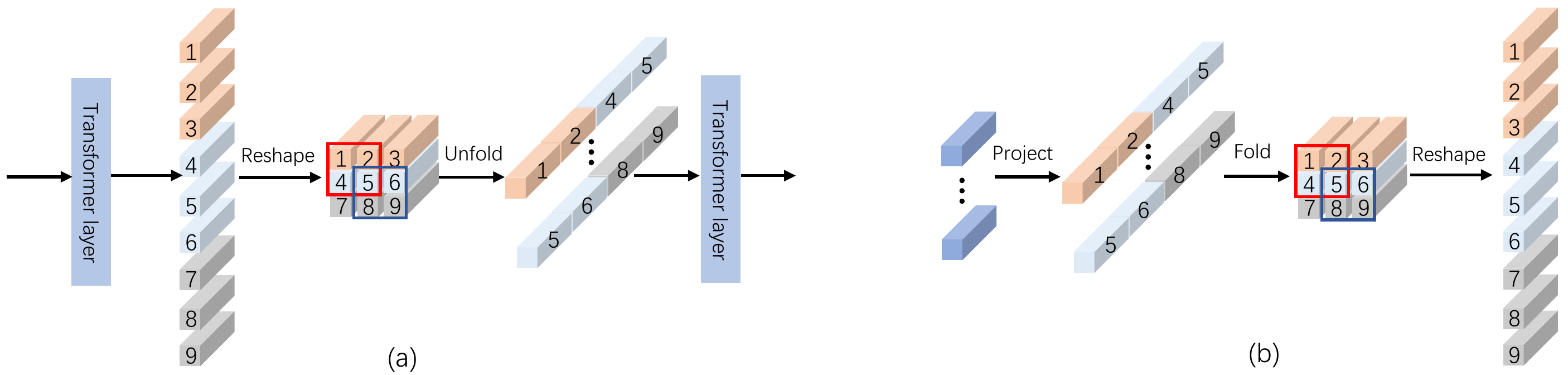}
  \caption{(a) T2T module merges neighbouring tokens into a new token, thus reducing the length of tokens. (b) Our proposed reverse T2T module upsamples tokens by expanding each token into multiple sub-tokens.}
  \label{T2T}
  \vspace{-0.3cm}
\end{figure*}

\vspace{-3mm}
\subsubsection{Tokens to Token}
Given a sequence of patch tokens $\bm{T}'$ with length $l$ from the previous layer, T2T-ViT iteratively applies the T2T module, which is composed of a re-structurization step and a soft split step, to model the local structure information in $\bm{T}'$ and obtain a new sequence of tokens.

\vspace{-3mm}
\paragraph{Re-structurization.}
As shown in Figure~\ref{T2T}(a), the tokens $\bm{T}'$ is first transformed using a transformer layer to obtain new tokens $\bm{T} \in {\mathbb{R}^{l\times c}}$:
\begin{equation} \label{reconstruction}
\bm{T} = \text{MLP}(\text{MSA}(\bm{T}')),
\end{equation}
where MSA and MLP denote the multi-head self-attention and multilayer perceptron in the original Transformer \cite{vaswani2017attention}, respectively. Note that layer normalization \cite{ba2016layer} is applied before each block.
Then, $\bm{T}$ is reshaped to a 2D image $\bm{I}\in{\mathbb{R}^{h\times{w\times c}}}$, where $l=h\times w$, to recover spatial structures, as shown in Figure~\ref{T2T}(a).

\vspace{-3mm}
\paragraph{Soft split.}
After the re-structurization step, $\bm{I}$ is first split into $k \times k$ patches with $s$ overlapping. $p$ zero-padding is also utilized to pad image boundaries. Then, the image patches are unfolded to a sequence of tokens $\bm{T}_o \in {\mathbb{R}^{l_o\times ck^2}}$, where the sequence length $l_o$ is computed as:
\begin{equation} \label{softsplit}
l_o =h_o\times w_o=\lfloor \frac{h+2p-k}{k-s}+1\rfloor \times \lfloor \frac{w+2p-k}{k-s}+1\rfloor.
\end{equation}
Different from ViT \cite{dosovitskiy2020image}, the overlapped patch splitting adopted in T2T-ViT introduces local correspondence within neighbouring patches, thus bringing spatial priors.

The T2T transformation can be conducted iteratively multiple times. In each time, the re-structurization step first transforms previous token embeddings to new embeddings and also integrates long-range dependencies within all tokens. Then, the soft split operation aggregates the tokens in each $k\times k$ neighbour into a new token, which is ready to use for the next layer. Furthermore, when setting $s<k-1$, the length of tokens can be reduced progressively.

We follow \cite{yuan2021tokens} to first soft split the input image into patches and then adopt the T2T module twice. Among the three soft split steps, the patch sizes are set to $k = [7, 3, 3]$, the overlappings are set to $s = [3, 1, 1]$, and the padding sizes are set to $p = [2, 1, 1]$.
As such, we can obtain multi-level tokens $\bm{T}_1 \in {\mathbb{R}^{l_1\times c}}$, $\bm{T}_2 \in {\mathbb{R}^{l_2\times c}}$, and $\bm{T}_3 \in {\mathbb{R}^{l_3\times c}}$.
Given the width and height of the input image as $H$ and $W$, respectively, then $l_1 = \frac{H}{4} \times \frac{W}{4}$, $l_2 = \frac{H}{8} \times \frac{W}{8}$, and $l_3 = \frac{H}{16} \times \frac{W}{16}$. We follow \cite{yuan2021tokens} to set $c=64$ and use a linear projection layer on $\bm{T}_3$ to transform its embedding dimension from $c$ to $d=384$.
\vspace{-3mm}
\subsubsection{Encoder with T2T-ViT Backbone}
The final token sequence $\bm{T}_3$ is added with the sinusoidal position embedding \cite{vaswani2017attention} to encode 2D position information. Then, $L^{\mathcal{E}}$ transformer layers are used to model long-range dependencies among $\bm{T}_3$ to extract powerful patch token embeddings $\bm{T}^{\mathcal{E}} \in {\mathbb{R}^{l_3\times d}}$.

For RGB SOD, we adopt a single transformer encoder to obtain RGB encoder patch tokens $\bm{T}_r^{\mathcal{E}} \in {\mathbb{R}^{l_3\times d}}$ from each input RGB image. For RGB-D SOD, we follow two-stream architectures to further use another transformer encoder to extract the depth encoder patch tokens $\bm{T}_d^{\mathcal{E}}$ from the input depth map in a similar way, as shown in Figure~\ref{VST}.

%%%%%%%%%%%%%%%%%%%%%%%%%%%%%%%%%%%%%%%%%%%%%%%%%%%%%%%%%%%%%%%%%%%%%%%%%%%%
\subsection{Transformer Convertor}
We insert a convertor module between the transformer encoder and decoder to convert the encoder patch tokens $\bm{T}_*^{\mathcal{E}}$ from the encoder space to the decoder space, thus obtaining the converted patch tokens $\bm{T}^{\mathcal{C}} \in {\mathbb{R}^{l_3 \times d}}$.
\vspace{-3mm}
\subsubsection{RGB-D Convertor}
We fuse $\bm{T}_r^{\mathcal{E}}$ and $\bm{T}_d^{\mathcal{E}}$ in the RGB-D converter to integrate the complementary information between the RGB and depth data.
To this end, we design a Cross Modality Transformer (CMT), which consists of $L^{\mathcal{C}}$ alternating cross-modality-attention layers and self-attention layers.

\vspace{-3mm}
\paragraph{Cross-modality-attention.}
Under the pure transformer architecture, we modify the standard self-attention layer to propagate long-range cross-modal dependencies between the image and depth data, thus obtaining the cross-modality-attention, which is detailed as follows.

First, similar with the self-attention in \cite{vaswani2017attention}, $\bm{T}_r^{\mathcal{E}}$ is embedded to queries $\bm{Q}_r \in {\mathbb{R}^{l_3\times d}}$, keys $\bm{K}_r \in {\mathbb{R}^{l_3\times d}}$, and values $\bm{V}_r \in {\mathbb{R}^{l_3\times d}}$ through three linear projections.
Similarly, we can obtain the depth queries $\bm{Q}_d$, keys $\bm{K}_d$, and values $\bm{V}_d$ from $\bm{T}_d^{\mathcal{E}}$.

Next, we compute the ``Scaled Dot-Product Attention" \cite{vaswani2017attention} between the queries from one modality with the keys from the other modality. Then, the output is computed as a weighted sum of the values, formulated as:
\begin{equation} \label{cross-at}
\begin{split}
\text{Attention}(\bm{Q}_r, \bm{K}_d, \bm{V}_d) = \text{softmax}(\bm{Q}_r\bm{K}_d^{\top} / \sqrt{d})\bm{V}_d, \\
\text{Attention}(\bm{Q}_d, \bm{K}_r, \bm{V}_r) = \text{softmax}(\bm{Q}_d\bm{K}_r^{\top} / \sqrt{d})\bm{V}_r.
\end{split}
\end{equation}

We follow the standard Transformer architecture in \cite{vaswani2017attention} and adopt the multi-head attention mechanism in the cross-modality-attention. The same positionwise feed-forward network, residual connections, and layer normalization \cite{ba2016layer} are also used, forming our CMT layer.

After each adoption of the proposed CMT layer, we use one standard transformer layer 
on each RGB and depth patch token sequence, further enhancing their token embeddings. After alternately using CMT and transformer for $L^{\mathcal{C}}$ times, we fuse the obtained RGB tokens and depth tokens by concatenation and then project them to the final converted tokens $\bm{T}^{\mathcal{C}}$, as shown in Figure~\ref{VST}.

\vspace{-3mm}
\subsubsection{RGB Convertor}
To align with our RGB-D SOD model, for RGB SOD, we simply use $L^{\mathcal{C}}$ standard transformer layers on $\bm{T}_r^{\mathcal{E}}$ to obtain the converted patch token sequence $\bm{T}^{\mathcal{C}}$.

%%%%%%%%%%%%%%%%%%%%%%%%%%%%%%%%%%%%%%%%%%%%%%%%%%%%%%%%%%%%%%%%%%%%%%%%%%%%
\subsection{Multi-task Transformer Decoder}
Our decoder aims to decode the patch tokens $\bm{T}^{\mathcal{C}}$ to saliency maps. 
Hence,
we propose a novel token upsampling method with multi-level token fusion and a token-based multi-task decoder.
\subsubsection{Token Upsampling and Multi-level Token Fusion}
We argue that directly predicting saliency maps from $\bm{T}^{\mathcal{C}}$ can not obtain high-quality results since the length of $\bm{T}^{\mathcal{C}}$ is relatively small, \ie, $l_3 = \frac{H}{16} \times \frac{W}{16}$, which is limited for dense prediction.
Thus, we propose to upsample patch tokens first and then conduct dense prediction.
Most CNN-based methods \cite{GateNet,zhao2019EGNet,liu2020S2MA,Fu2020JLDCF} adopt bilinear upsampling to recover large scale feature maps. 
Alternatively, we propose a new token upsampling method under the transformer framework. Inspired by the T2T module \cite{yuan2021tokens} that aggregates neighbour tokens to reduce the length of tokens progressively, we propose a reverse T2T (RT2T) transformation to upsample tokens by expanding each token into multiple sub-tokens, as shown in Figure~\ref{T2T}(b).

Specifically, we first project the input patch tokens to reduce their embedding dimension from $d=384$ to $c=64$.
Then, we use another linear projection to expand the embedding dimension from $c$ to $c k^2$.
Next, similar to the soft split step in T2T, each token is seen as a $k\times k$ image patch and neighbouring patches have $s$ overlapping. Then, we can fold the tokens as an image using $p$ zero-padding. The output image size can be computed using \eqref{softsplit} reversely, \ie, given the length of the input patch tokens as $h_o\times w_o$, the spatial size of the out image is $h\times w$. Finally, we reshape the image back to the upsampled tokens with size $l_o \times c$, where $l_o=h\times w$.
By setting $s<k-1$, the RT2T transformation can increase the length of the tokens. Motivated by T2T-ViT, we use RT2T three times and set $k = [3, 3, 7]$, $s = [1, 1, 3]$, and $p = [1, 1, 3]$. Thus, the length of the patch tokens can be gradually upsampled to $H\times W$, equaling to the original size of the input image.

Furthermore, motivated by the widely proved successes of multi-level feature fusion in existing SOD methods \cite{hou2018dss,MINet-CVPR2020,GateNet,fan2020bbsnet,luo2020Cas-Gnn}, we leverage low-level tokens with larger lengths from the T2T-ViT encoder, \ie, $\bm{T}_1$ and $\bm{T}_2$, to provide accurate local structural information.
For both RGB and RGB-D SOD, we only use the low-level tokens from the RGB transformer encoder.
Concretely, we progressively fuse $\bm{T}_2$ and $\bm{T}_1$ with the upsampled patch tokens via concatenation and linear projection. Then, we adopt one transformer layer to obtain the decoder tokens $\bm{T}^{\mathcal{D}}_i$ at each level $i$, where $i=2,1$. The whole process is formulated as:
\begin{equation} \label{decoder}
\bm{T}^{\mathcal{D}}_{i} = \text{MLP}(\text{MSA}(\text{Linear}([\text{RT2T}(\bm{T}^{\mathcal{D}}_{i+1}),\bm{T}_{i}])),
\end{equation}
where $[,]$ means concatenation along the token embedding dimension. ``Linear" means linear projection to reduce the embedding dimension after the concatenation to $c$. Finally, we use another linear projection to recover the embedding dimension of $\bm{T}^{\mathcal{D}}_{i}$ back to $d$.

\subsubsection{Token Based Multi-task Prediction}
Inspired by existing pure transformer methods \cite{yuan2021tokens,dosovitskiy2020image}, which add a class token on the patch token sequence for image classification, we also leverage task-related tokens to predict results.
However, we can not obtain dense prediction results by directly using MLP on the task token embedding, as done in \cite{yuan2021tokens,dosovitskiy2020image}. Hence, we propose to perform patch-task-attention between the patch tokens and the task-related token to perform SOD.

In addition, motivated by the widely used boundary detection in SOD models \cite{zhao2019EGNet,CVPR2020_LDF,zhang2020select,Wei2020CoNet}, we also adopt the multi-task learning strategy to jointly perform saliency and boundary detection, thus using the latter to help boost the performance of the former.

To this end, we design two task-related tokens, \ie, a saliency token $\bm{t}_s\in{\mathbb{R}^{1 \times d}}$ and a boundary token $\bm{t}_b\in{\mathbb{R}^{1 \times d}}$.
At each decoder level $i$, we add the saliency and boundary tokens $\bm{t}_s$ and $\bm{t}_b$ on the patch token sequence $\bm{T}^{\mathcal{D}}_{i}$, and then process them using $L^{\mathcal{D}}_i$ transformer layers. As such, the two task tokens can learn image-dependent task-related embeddings from the interaction with the patch tokens. After this, we take the updated patch tokens as input and perform the token upsampling and multi-level fusion process in \eqref{decoder} to obtain upsampled patch tokens $\bm{T}^{\mathcal{D}}_{i-1}$. Next, we reuse the updated $\bm{t}_s$ and $\bm{t}_b$ in the next level $i-1$ to further update them and $\bm{T}^{\mathcal{D}}_{i-1}$. We repeat this process until we reach the last decoder level with the $\frac{1}{4}$ scale.

For saliency and boundary prediction, we perform patch-task-attention between the final decoder patch tokens $\bm{T}^{\mathcal{D}}_{1}$ and the saliency and boundary tokens $\bm{t}_s$ and $\bm{t}_b$.
For saliency prediction, we first embed $\bm{T}^{\mathcal{D}}_{1}$ to queries $\bm{Q}^{\mathcal{D}}_s \in {\mathbb{R}^{l_1\times d}}$ and embed $\bm{t}_s$ to a key $\bm{K}_s \in {\mathbb{R}^{1\times d}}$ and a value $\bm{V}_s \in {\mathbb{R}^{1\times d}}$.
Similarly, for boundary prediction, we embed $\bm{T}^{\mathcal{D}}_{1}$ to $\bm{Q}^{\mathcal{D}}_b$ and embed $\bm{t}_b$ to $\bm{K}_b$ and $\bm{V}_b$.
Then, we adopt the patch-task-attention to obtain the task-related patch tokens:
\begin{equation} \label{predict}
\begin{split}
\bm{T}_s^{\mathcal{D}} = \text{sigmoid}(\bm{Q}_s^{\mathcal{D}}\bm{K}_s^{\top} / \sqrt{d})\bm{V}_s + \bm{T}^{\mathcal{D}}_{1}, \\
\bm{T}_b^{\mathcal{D}} = \text{sigmoid}(\bm{Q}_b^{\mathcal{D}}\bm{K}_b^{\top} / \sqrt{d})\bm{V}_b + \bm{T}^{\mathcal{D}}_{1}.
\end{split}
\end{equation}
Here we use the sigmoid activation for the attention computation since in each equation we only have one key.

%-------------------------------------------------------------------------
\begin{table*}[t]
\centering
\scriptsize
\renewcommand{\arraystretch}{1.0}
\renewcommand{\tabcolsep}{1.0mm}
\caption{Ablation studies of our proposed model. ``Bili'' denotes bilinear upsampling. ``F" means multi-level token fusion. ``TMD" denotes our proposed token-based multi-task decoder, while ``C2D'' means using conventional two-stream decoder to perform saliency and boundary detection without using task-related tokens. The best results are labeled in \blu{blue}.
}
\begin{tabular}{l|l|cccc|cccc|cccc|cccc}
\hline
\multicolumn{2}{l|}{\multirow{2}{*}{Settings}} & \multicolumn{4}{c|}{NJUD \cite{ju2014njud}} & \multicolumn{4}{c|}{DUTLF-Depth \cite{Piao2019dmra}} & \multicolumn{4}{c|}{STERE \cite{niu2012stere}} & \multicolumn{4}{c}{LFSD \cite{li2014lfsd}}\\
\multicolumn{2}{l|}{} & \multicolumn{1}{l}{$S_m \uparrow$} & \multicolumn{1}{l}{maxF $\uparrow$} & \multicolumn{1}{l}{$E_{\xi}^{\text{max}} \uparrow$} & \multicolumn{1}{l|}{MAE $\downarrow$}
                      & \multicolumn{1}{l}{$S_m \uparrow$} & \multicolumn{1}{l}{maxF $\uparrow$} & \multicolumn{1}{l}{$E_{\xi}^{\text{max}} \uparrow$} & \multicolumn{1}{l|}{MAE $\downarrow$}
                      & \multicolumn{1}{l}{$S_m \uparrow$} & \multicolumn{1}{l}{maxF $\uparrow$} & \multicolumn{1}{l}{$E_{\xi}^{\text{max}} \uparrow$} & \multicolumn{1}{l|}{MAE $\downarrow$}
                      & \multicolumn{1}{l}{$S_m \uparrow$} & \multicolumn{1}{l}{maxF $\uparrow$} & \multicolumn{1}{l}{$E_{\xi}^{\text{max}} \uparrow$} & \multicolumn{1}{l}{MAE $\downarrow$}
  \\ \hline

\multicolumn{2}{l|}{Baseline}       &0.869 &0.862 &0.931 &0.073   &0.889 &0.887 &0.942 &0.062   &0.868 &0.853 &0.927 &0.075   &0.842 &0.845 &0.893 &0.103\\ \hline
\multicolumn{2}{l|}{+CMT}           &0.873 &0.867 &0.934 &0.072   &0.889 &0.890 &0.942 &0.063   &0.869 &0.854 &0.928 &0.075   &0.849 &0.855 &0.900 &0.100\\ \hline
\multicolumn{2}{l|}{+CMT+Bili}      &0.906 &0.902 &0.944 &0.045   &0.926 &0.930 &0.961 &0.032   &0.889 &0.877 &0.939 &0.051   &0.856 &0.858 &0.895 &0.081\\
\multicolumn{2}{l|}{+CMT+RT2T}       &0.915 &0.915 &0.951 &0.039   &0.934 &0.940 &0.964 &0.028   &0.896 &0.889 &0.943 &0.046   &0.867 &0.873 &0.903 &0.073\\ \hline
\multicolumn{2}{l|}{+CMT+RT2T+F}    &\blu{0.923} &\blu{0.923} &\blu{0.954} &\blu{0.035}   &0.936 &0.943 &0.963 &0.028   &0.910 &0.903 &0.947 &0.040   &0.876 &0.880 &0.909 &0.067\\
\multicolumn{2}{l|}{+CMT+RT2T+F+TMD}  &0.922 &0.920 &0.951 &\blu{0.035}   &\blu{0.943} &\blu{0.948} &\blu{0.969} &\blu{0.024}   &\blu{0.913} &\blu{0.907} &\blu{0.951} &\blu{0.038}   &\blu{0.882} &\blu{0.889} &\blu{0.921} &\blu{0.061}\\ \hline
\multicolumn{2}{l|}{+CMT+RT2T+F+C2D}  &0.922 &0.921 &\blu{0.954} &0.036   &0.941 &0.947 &0.968 &0.026   &0.911 &0.906 &0.949 &0.040   &0.874 &0.878 &0.909 &0.069\\ \hline
\end{tabular}
\label{ablationTab}
\vspace{-3mm}
\end{table*}

Since $\bm{T}_s^{\mathcal{D}}$ and $\bm{T}_b^{\mathcal{D}}$ are at the $\frac{1}{4}$ scale, we adopt the third RT2T transformation to upsample them to the full resolution.  Finally, we apply two linear transformations with the sigmoid activation to project them to scalars in $[0,1]$, and then reshape them to a 2D saliency map and a 2D boundary map, respectively. The whole process is given in Figure~\ref{VST}.

\section{Experiments}
\subsection{Datasets and Evaluation Metrics}
For RGB SOD, we evaluate our VST model on six widely used benchmark datasets, including \textbf{ECSSD} \cite{yan2013ECSSD} (1,000 images), \textbf{HKU-IS} \cite{li2015HKUIS} (4,447 images), \textbf{PASCAL-S} \cite{li2014PASCALS} (850 images), \textbf{DUT-O} \cite{yang2013DUTO} (5,168 images), \textbf{SOD} \cite{movahedi2010SOD} (300 images), and \textbf{DUTS} \cite{wang2017duts} (10,553 training images and 5,019 testing images).
For RGB-D SOD, we use nine widely used benchmark datasets: \textbf{STERE} \cite{niu2012stere} (1,000 image pairs), \textbf{LFSD} \cite{li2014lfsd} (100 image pairs), \textbf{RGBD135} \cite{cheng2014rgbd135} (135 image pairs), \textbf{SSD} \cite{zhu2017ssd} (80 image pairs), \textbf{NJUD} \cite{ju2014njud} (1,985 image pairs), \textbf{NLPR} \cite{peng2014nlpr} (1,000 image pairs), \textbf{DUTLF-Depth} \cite{Piao2019dmra} (1,200 image pairs), \textbf{SIP} \cite{fan2020SIP} (929 image pairs), and \textbf{ReDWeb-S} \cite{liu2020ReDWeb-S} (3,179 image pairs).

We adopt four widely used evaluation metrics to evaluate our model performance comprehensively. Specifically, 
Structure-measure $S_m$ \cite{fan2017structure} evaluates region-aware and object-aware structural similarity. Maximum F-measure (maxF) jointly considers precision and recall under the optimal threshold. Maximum enhanced-alignment measure $E_{\xi}^{\text{max}}$ \cite{Fan2018Enhanced} simultaneously considers pixel-level errors and image-level errors. Mean Absolute Error (MAE) computes pixel-wise average absolute error. To evaluate the model complexity, we also report the multiply accumulate operations (MACs) and the number of parameters (Params).

\subsection{Implementation Details}
For fair comparisons,
we follow most previous methods to use the training set of DUTS to train our VST for RGB SOD and use 1,485 images from NJUD, 700 images from NLPR, and 800 images from DUTLF-Depth to train our VST for RGB-D SOD.
We follow \cite{zhao2019EGNet} to use a sober operator to generate the boundary ground truth from GT saliency maps.
For depth data preprocessing, we normalize the depth maps to [0,1] and duplicate them to three channels.
Finally, we resize each image or depth map to $256 \times 256$ pixels and then randomly crop $224 \times 224$ image regions as the model input and use random flipping as data augmentation.

We use the pre-trained T2T-ViT$_t$-14 \cite{yuan2021tokens} model as our backbone since it has similar computational complexity as ResNet50 \cite{he2016resnet} does. This model uses the efficient Performer \cite{2020performer} and $c=64$ in T2T modules, and sets $L^{\mathcal{E}}=14$. In our convertor and decoder, we set $L^{\mathcal{C}}=L^{\mathcal{D}}_3=4$ and $L^{\mathcal{D}}_2=L^{\mathcal{D}}_1=2$ according to experimental results.
We set the batchsizes as 11 and 8, and the total training steps as 40,000 and 60,000, for RGB and RGB-D SOD, respectively.
For both of them, Adam \cite{Adam2015} is adopted as the optimizer and the binary cross entropy loss is used for both saliency and boundary prediction.
The initial learning rate is set to 0.0001 and reduced by a factor of 10 at half and three-quarters of the total step, respectively.
Deep supervision is also used to facilitate the model training, where we use the patch-task attention to predict saliency and boundary at each decoder level.
We implemented our model using Pytorch \cite{paszke2019pytorch} and trained it on a GTX 1080 Ti GPU.

\begin{table*}[t]
  \centering
  \footnotesize
  \renewcommand{\arraystretch}{1.0}
  \renewcommand{\tabcolsep}{0.7mm}
 \caption{Quantitative comparison of our proposed VST with other 14 SOTA RGB-D SOD methods on 9 benchmark datasets. \red{Red} and \blu{blue} denote the best and the second-best results, respectively. `-' indicates the code or result is not available.}
  \begin{tabular}{lr|cccccccccccccc|c}
  \hline

    Dataset
    & Metric
    & A2dele &JL-DCF & SSF-RGBD & UC-Net & $S^2$MA & PGAR & DANet & cmMS & ATST  & CMW & Cas-Gnn & HDFNet & CoNet & BBS-Net & VST\\
    &
    & \cite{piao2020a2dele} &\cite{Fu2020JLDCF} & \cite{zhang2020select}   & \cite{zhang2020ucnet} & \cite{liu2020S2MA} & \cite{chen2020PGAR} & \cite{zhao2020DANet} & \cite{li2020cmMS} & \cite{zhang2020ATSA} &\cite{Li2020CMWNet} & \cite{luo2020Cas-Gnn} & \cite{HDFNet-ECCV2020} & \cite{Wei2020CoNet} & \cite{fan2020bbsnet}\\ \hline

   \multicolumn{2}{r|}{MACs (G)}  &41.86 &211.06 &46.56 &16.16 &141.19	&44.65 &66.25 &134.77 &42.17 &208.03 &-	&91.77 &20.89 &31.2 &30.99\\
   \multicolumn{2}{r|}{Params (M)} &30.34 &143.52 &32.93 &31.26 &86.65	&16.2 &26.68 &92.02 &32.17 &85.65 &- &44.15	&43.66 &49.77 &83.83\\
     \hline
     
  \multirow{4}{*}{NJUD}
    & $S_m\uparrow$   &0.871 &0.902 &0.899 &0.897 &0.894 &0.909 &0.899 &0.900 &0.885 &0.870 &0.911 &0.908 &0.896 &\blu{0.921} &\red{0.922}\\
    & maxF$\uparrow$  &0.874 &0.904 &0.896 &0.895 &0.889 &0.907 &0.898 &0.897 &0.893 &0.871 &0.916 &0.911 &0.893 &\blu{0.919} &\red{0.920} \\
    & $E_{\xi}^{\text{max}}\uparrow$ &0.916 &0.944 &0.935 &0.936 &0.930 &0.940 &0.935 &0.936 &0.930 &0.927 &0.948 &0.944 &0.937 &\blu{0.949} &\red{0.951}\\
  \cite{ju2014njud}& MAE$\downarrow$ &0.051 &0.041 &0.043 &0.043 &0.054 &0.042 &0.046 &0.044 &0.047 &0.061 &\blu{0.036} &0.039 &0.046 &\red{0.035} &\red{0.035} \\
     \hline
  \multirow{4}{*}{NLPR}
    & $S_m\uparrow$   &0.899 &0.925 &0.915 &0.920 &0.916 &0.917 &0.920 &0.919 &0.909 &0.917 &0.919 &0.923 &0.912 &\blu{0.931} &\red{0.932}\\
    & maxF$\uparrow$  &0.882 &\blu{0.918} &0.896 &0.903 &0.902 &0.897 &0.909 &0.904 &0.898 &0.903 &0.906 &0.917 &0.893 &\blu{0.918} &\red{0.920}\\
    & $E_{\xi}^{\text{max}}\uparrow$ &0.944 &\red{0.963} &0.953 &0.956 &0.953 &0.950 &0.955 &0.955 &0.951 &0.951 &0.955 &\red{0.963} &0.948 &0.961 &\blu{0.962}\\
   \cite{peng2014nlpr}& MAE$\downarrow$ &0.029 &\red{0.022} &0.027 &0.025 &0.030 &0.027 &0.027 &0.028 &0.027 &0.029 &0.025 &\blu{0.023} &0.027 &0.023 &0.024\\
     \hline
  \multirow{4}{*}{}
    & $S_m\uparrow$  &0.885 &0.906 &0.915 &0.871 &0.904 &0.899 &0.899 &0.912 &0.916 &0.797 &0.920 &0.908 &\blu{0.923} &0.882 &\red{0.943}\\
    DUTLF& maxF$\uparrow$ &0.891 &0.910 &0.923 &0.864 &0.899 &0.898 &0.904 &0.913 &0.928 &0.779 &0.926 &0.915 &\blu{0.932} &0.870 &\red{0.948}\\
    -Depth& $E_{\xi}^{\text{max}}\uparrow$  &0.928 &0.941 &0.950 &0.908 &0.935 &0.933 &0.939 &0.940 &0.953 &0.864 &0.953 &0.945 &\blu{0.959} &0.912 &\red{0.969}\\
     \cite{Piao2019dmra}& MAE$\downarrow$ &0.043 &0.042 &0.033 &0.059 &0.043 &0.041 &0.042 &0.036 &0.033 &0.098 &0.030 &0.041 &\blu{0.029} &0.058 &\red{0.024}\\
     \hline
  \multirow{4}{*}{ReDWeb-S}
    & $S_m\uparrow$   &0.641 &\blu{0.734} &0.595 &0.713 &0.711 &0.656 &-	 &0.699 &0.679 &0.634 &- &0.728 &0.696 &0.693 &\red{0.759}\\
    & maxF$\uparrow$   &0.603 &\blu{0.727} &0.558 &0.710 &0.696 &0.632 &- &0.677 &0.673 &0.607 &- &0.717 &0.693 &0.680 &\red{0.763}\\
    & $E_{\xi}^{\text{max}}\uparrow$  &0.674 &\blu{0.805} &0.710 &0.794 &0.781 &0.749 &- &0.767 &0.758 &0.714 &- &0.804 &0.782 &0.763 &\red{0.826}\\
    \cite{liu2020ReDWeb-S}& MAE$\downarrow$  &0.160 &\blu{0.128} &0.189 &0.130 &0.139 &0.161 &- &0.143 &0.155 &0.195 &- &0.129 &0.147 &0.150 &\red{0.113}\\
     \hline

  \multirow{4}{*}{STERE}
    & $S_m\uparrow$   &0.879 &0.903 &0.837 &0.903 &0.890 &0.894 &0.901 &0.894 &0.896 &0.852 &0.899 &0.900 &0.905 &\blu{0.908} &\red{0.913} \\
    & maxF$\uparrow$  &0.880 &\blu{0.904} &0.840 &0.899 &0.882 &0.880 &0.892 &0.887 &0.901 &0.837 &0.901 &0.900 &0.901 &0.903 &\red{0.907} \\
    & $E_{\xi}^{\text{max}}\uparrow$ &0.928 &\blu{0.947} &0.912 &0.944 &0.932 &0.929 &0.937 &0.935 &0.942 &0.907 &0.944 &0.943 &\blu{0.947} &0.942 &\red{0.951} \\
    \cite{niu2012stere}  & MAE$\downarrow$ &0.045 &0.040 &0.065 &0.039 &0.051 &0.045 &0.044 &0.045 &\blu{0.038} &0.067 &0.039 &0.042 &\red{0.037} &0.041 &\blu{0.038}\\
     \hline

  \multirow{4}{*}{SSD}
    & $S_m\uparrow$   &0.803 &0.860 &0.790 &0.865 &0.868 &0.832 &0.864 &0.857 &0.850 &0.798 &0.872 &\blu{0.879} &0.851 &0.863 &\red{0.889}\\
    & maxF$\uparrow$  &0.777 &0.833 &0.762 &0.855 &0.848 &0.798 &0.843 &0.839 &0.853 &0.771 &0.863 &\blu{0.870} &0.837 &0.843 &\red{0.876} \\
    & $E_{\xi}^{\text{max}}\uparrow$  &0.862 &0.902 &0.867 &0.907 &0.909 &0.872 &0.914 &0.900 &0.920 &0.871 &0.923 &\blu{0.925} &0.917 &0.914 &\red{0.935}\\
    \cite{zhu2017ssd}& MAE$\downarrow$ &0.070 &0.053 &0.084 &0.049 &0.053 &0.068 &0.050 &0.053 &0.052 &0.085 &0.047 &\blu{0.046} &0.056 &0.052 &\red{0.045}\\
     \hline

  \multirow{4}{*}{RGBD135}
    & $S_m\uparrow$   &0.886 &0.931 &0.904 &0.934 &\blu{0.941} &0.886 &0.924 &0.934 &0.917 &0.934 &0.894 &0.926 &0.914 &0.934 &\red{0.943}\\
    & maxF$\uparrow$  &0.872 &0.923 &0.885 &0.930 &\blu{0.935} &0.864 &0.914 &0.928 &0.916 &0.931 &0.894 &0.921 &0.902 &0.928 &\red{0.940} \\
    & $E_{\xi}^{\text{max}}\uparrow$ &0.921 &0.968 &0.940 &\blu{0.976} &0.973 &0.924 &0.966 &0.969 &0.961 &0.969 &0.937 &0.970 &0.948 &0.966 &\red{0.978} \\
    \cite{cheng2014rgbd135}& MAE$\downarrow$ &0.029 &0.021 &0.026 &0.019 &0.021 &0.032 &0.023 &\blu{0.018} &0.022 &0.022 &0.028 &0.022 &0.024 &0.021 &\red{0.017} \\
     \hline

  \multirow{4}{*}{LFSD}
    & $S_m\uparrow$  &0.825 &0.853 &0.851 &\blu{0.856} &0.829 &0.808 &0.841 &0.845 &0.845 &0.776 &0.838 &0.846 &0.848 &0.835 &\red{0.882} \\
    & maxF$\uparrow$  &0.828 &\blu{0.863} &\blu{0.863} &0.860 &0.831 &0.794 &0.840 &0.858 &0.859 &0.779 &0.843 &0.858 &0.852 &0.828 &\red{0.889} \\
    & $E_{\xi}^{\text{max}}\uparrow$  &0.866 &0.894 &0.892 &\blu{0.898} &0.865 &0.853 &0.874 &0.886 &0.893 &0.834 &0.880 &0.889 &0.895 &0.870 &\red{0.921}\\
    \cite{li2014lfsd}& MAE$\downarrow$  &0.084 &0.077 &\blu{0.074} &\blu{0.074} &0.102 &0.099 &0.087 &0.082 &0.078 &0.130 &0.081 &0.085 &0.076 &0.092 &\red{0.061} \\
     \hline

  \multirow{4}{*}{SIP}
    & $S_m\uparrow$  &0.829 &0.880 &0.799 &0.875 &0.872 &0.838 &0.875 &0.872 &0.849 &0.705 &- &\blu{0.886} &0.860 &0.879 &\red{0.904}\\
    & maxF$\uparrow$  &0.834 &0.889 &0.786 &0.879 &0.877 &0.827 &0.876 &0.876 &0.861 &0.677 &- &\blu{0.894} &0.873 &0.884 &\red{0.915}\\
    & $E_{\xi}^{\text{max}}\uparrow$  &0.890 &0.925 &0.870 &0.919 &0.919 &0.886 &0.918 &0.911 &0.901 &0.804 &- &\blu{0.930} &0.917 &0.922 &\red{0.944}\\
    \cite{fan2020SIP} & MAE$\downarrow$ &0.070 &0.049 &0.091 &0.051 &0.058 &0.073 &0.055 &0.058 &0.063 &0.141 &- &\blu{0.048} &0.058 &0.055 &\red{0.040}\\
     \hline

  \end{tabular}
  \label{RGBD_SOTA}
   \vspace{-4mm}
\end{table*}

\subsection{Ablation Study}\label{sec:ablation}
Since our RGB-D VST is built by adding one more transformer encoder and additional CMT based on our RGB VST, while the other parts of the two models are the same, we conduct ablation studies based on our RGB-D VST to verify all of our proposed model components.
The experimental results on four RGB-D SOD datasets, \ie, NJUD, DUTLF-Depth, STERE, and LFSD, are given in Table~\ref{ablationTab}.
We remove the transformer convertor and the decoder from our RGB-D VST as the baseline model. 
Specifically, it uses the two-stream transformer encoder to extract RGB encoder patch tokens $\bm{T}_r^{\mathcal{E}}$ and the depth encoder patch tokens $\bm{T}_d^{\mathcal{E}}$, and then directly concatenate them and predict the saliency map with 1/16 scale by using MLP on each patch token.

\vspace{-3mm}
\paragraph{Effectiveness of CMT.}
For cross-modal information fusion, we deploy our proposed CMT right after the transformer encoder to substitute the concatenation fusion method in the baseline model, shown as ``+CMT" in Table~\ref{ablationTab}.
Compared to the baseline, CMT brings performance gain especially on the NJUD and LFSD datasets, hence demonstrating its effectiveness.

\vspace{-3mm}
\paragraph{Effectiveness of RT2T.}
Based on ``+CMT'' model, we further simply use bilinear upsampling (``+CMT+Bili") to progressively upsample tokens to the full resolution and then predict the saliency map.
The results show using bilinear upsampling to increase the resolution of the saliency map can largely improve the model performance.
Then, we replace bilinear upsampling with our proposed RT2T token upsampling method (``+CMT+RT2T"). We find that RT2T leads to obvious performance improvement compared with using bilinear upsampling, which verifies its effectiveness. 

\vspace{-3mm}
\paragraph{Effectiveness of multi-level token fusion.}
We progressively fuse $\bm{T}_1$ and $\bm{T}_2$ in our decoder (``+CMT+RT2T+F") to supply low-level fine-grained information. We find that this strategy further improves the model performance. Hence, leveraging low-level tokens in transformer is as important as fusing low-level features in CNN-based models.

\begin{table*}[t]
  \centering
  \footnotesize
  \renewcommand{\arraystretch}{1.0}
  \renewcommand{\tabcolsep}{1mm}
 \caption{Quantitative comparison of our proposed VST with other 12 SOTA RGB SOD methods on 6 benchmark datasets. ``-R" and ``-R2" means the ResNet50 and Res2Net backbone, respectively.}
  \begin{tabular}{lr|cccccccccccc|c}
  \hline

    Dataset
    & Metric
    & PiCANet & BASNet & CPD-R & PoolNet  & AFNet & TSPOANet & EGNet-R & ITSD-R & MINet-R & LDF-R & CSF-R2 & GateNet-R & VST\\
    &
    & \cite{liu2018picanet} &\cite{Qin19BASNet} & \cite{Wu_CPD}   & \cite{Liu19PoolNet} & \cite{Feng_AFNet}& \cite{Liu_TSPOANet} & \cite{zhao2019EGNet} & \cite{Zhou2020ITSD} & \cite{MINet-CVPR2020} &\cite{CVPR2020_LDF} & \cite{gao2020sod100k} &\cite{GateNet}\\ \hline

   \multicolumn{2}{r|}{MACs (G)}  &54.05 &127.36 &17.77 &88.89	&21.66	&-	&157.21	&15.96 &87.11 &15.51 &18.96	&162.13	&23.16\\
   \multicolumn{2}{r|}{Params (M)}  &47.22	&87.06 &47.85 &68.26 &35.95	&- &111.64 &26.47 &162.38 &25.15 &36.53	&128.63	&44.48\\
     \hline
     
  \multirow{4}{*}{DUTS}
    & $S_m\uparrow$  &0.863 &0.866 &0.869 &0.879 &0.867 &0.860 &0.887 &0.885 &0.884 &\blu{0.892} &0.890 &0.891 &\red{0.896} \\
    & maxF$\uparrow$ &0.840 &0.838 &0.840 &0.853 &0.838 &0.828 &0.866 &0.867 &0.864 &\red{0.877} &0.869 &\blu{0.874} &\red{0.877}\\
    & $E_{\xi}^{\text{max}}\uparrow$ &0.915 &0.902 &0.913 &0.917 &0.910 &0.907 &0.926 &0.929 &0.926 &0.930 &0.929 &\blu{0.932} &\red{0.939}\\
  \cite{wang2017duts}& MAE$\downarrow$ &0.040 &0.047 &0.043 &0.041 &0.045 &0.049 &0.039 &0.041 &\blu{0.037} &\red{0.034} &\blu{0.037} &0.038 &\blu{0.037}\\
     \hline
  \multirow{4}{*}{ECSSD}
    & $S_m\uparrow$   &0.916 &0.916 &0.918 &0.917 &0.914 &0.907 &0.925 &0.925 &0.925 &0.925 &\blu{0.931} &0.924 &\red{0.932}\\
    & maxF$\uparrow$  &0.929 &0.931 &0.926 &0.929 &0.924 &0.919 &0.936 &0.939 &0.938 &0.938 &\blu{0.942} &0.935 &\red{0.944}\\
    & $E_{\xi}^{\text{max}}\uparrow$ &0.953 &0.951 &0.951 &0.948 &0.947 &0.942 &0.955 &0.959 &0.957 &0.954 &\blu{0.960} &0.955 &\red{0.964} \\
   \cite{yan2013ECSSD}& MAE$\downarrow$ &0.035 &0.037 &0.037 &0.042 &0.042 &0.047 &0.037 &0.035 &\blu{0.034} &\blu{0.034} &\red{0.033} &0.038 &\blu{0.034}\\
     \hline
  \multirow{4}{*}{HKU-IS}
    & $S_m\uparrow$  &0.905 &0.909 &0.906 &0.916 &0.905 &0.902 &0.918 &0.917 &0.919 &0.920 &- &\blu{0.921} &\red{0.928} \\
    & maxF$\uparrow$ &0.913 &0.919 &0.911 &0.920 &0.910 &0.909 &0.923 &0.926 &0.926 &\blu{0.929} &- &0.926 &\red{0.937} \\
    & $E_{\xi}^{\text{max}}\uparrow$ &0.951 &0.952 &0.950 &0.955 &0.949 &0.950 &0.956 &\blu{0.960} &\blu{0.960} &0.958 &- &0.959 &\red{0.968}\\
     \cite{li2015HKUIS}& MAE$\downarrow$ &0.031 &0.032 &0.034 &0.032 &0.036 &0.039 &0.031 &0.031 &\blu{0.029} &\red{0.028} &- &0.031 &0.030\\
     \hline
  \multirow{4}{*}{PASCAL-S}
    & $S_m\uparrow$   &0.846 &0.837 &0.847 &0.852 &0.849 &0.841 &0.852 &0.861 &0.856 &0.861 &\blu{0.863} &\blu{0.863} &\red{0.873}\\
    & maxF$\uparrow$  &0.824 &0.819 &0.817 &0.830 &0.824 &0.817 &0.825 &\blu{0.839} &0.831 &\blu{0.839} &\blu{0.839} &0.836 &\red{0.850}\\
    & $E_{\xi}^{\text{max}}\uparrow$  &0.882 &0.868 &0.872 &0.880 &0.877 &0.871 &0.874 &\blu{0.889} &0.883 &0.888 &0.885 &0.886 &\red{0.900} \\
    \cite{li2014PASCALS}& MAE$\downarrow$ &0.072 &0.083 &0.077 &0.076 &0.076 &0.082 &0.080 &\blu{0.071} &\blu{0.071} &\red{0.067} &0.073 &\blu{0.071} &\red{0.067}\\
     \hline

  \multirow{4}{*}{DUT-O}
    & $S_m\uparrow$  &0.826 &0.836 &0.825 &0.832 &0.826 &0.818 &\blu{0.841} &0.840 &0.833 &0.839 &0.838 &0.840 &\red{0.850} \\
    & maxF$\uparrow$  &0.767 &0.779 &0.754 &0.769 &0.759 &0.750 &0.778 &\blu{0.792} &0.769 &0.782 &0.775 &0.782 &\red{0.800}\\
    & $E_{\xi}^{\text{max}}\uparrow$ &0.865 &0.872 &0.868 &0.869 &0.861 &0.858 &0.878 &\blu{0.880} &0.869 &0.870 &0.869 &0.878 &\red{0.888}\\
    \cite{yang2013DUTO} & MAE$\downarrow$ &0.054 &0.057 &0.056 &0.056 &0.057 &0.062 &\blu{0.053} &0.061 &0.056 &\red{0.052} &0.055 &0.055 &0.058\\
     \hline

  \multirow{4}{*}{SOD}
    & $S_m\uparrow$   &0.813 &0.799 &0.797 &0.823 &0.811 &0.802 &0.824 &\blu{0.835} &0.830 &0.831 &0.826 &0.827 &\red{0.854} \\
    & maxF$\uparrow$  &0.824 &0.808 &0.804 &0.832 &0.819 &0.809 &0.831 &\blu{0.849} &0.835 &0.841 &0.832 &0.835 &\red{0.866} \\
    & $E_{\xi}^{\text{max}}\uparrow$  &0.871 &0.846 &0.860 &0.873 &0.867 &0.852 &0.875 &\blu{0.889} &0.878 &0.878 &0.883 &0.877 &\red{0.902}\\
    \cite{movahedi2010SOD}& MAE$\downarrow$  &0.073 &0.091 &0.089 &0.085 &0.085 &0.094 &0.080 &0.075 &0.074 &\blu{0.071} &0.079 &0.079 &\red{0.065}\\
     \hline

  \end{tabular}
  \label{RGB_SOTA}
\end{table*}

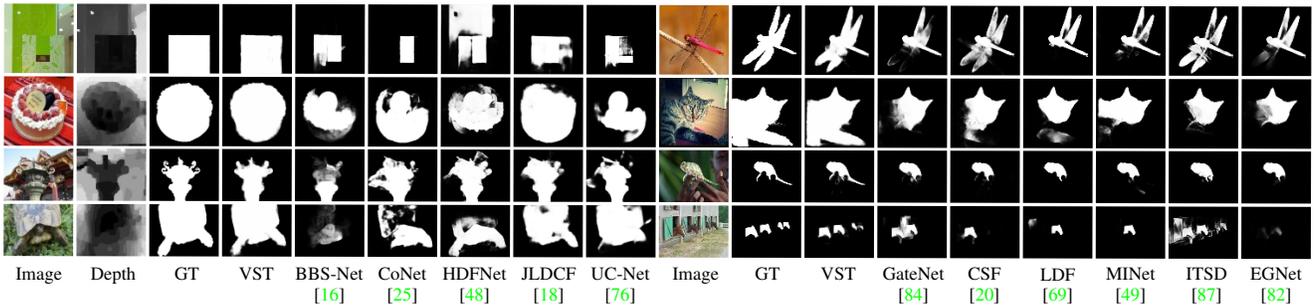
\begin{figure*}[t]
  \graphicspath{{Figures/qualitative/}}
  \centering
  \begin{overpic}[width=1\linewidth]{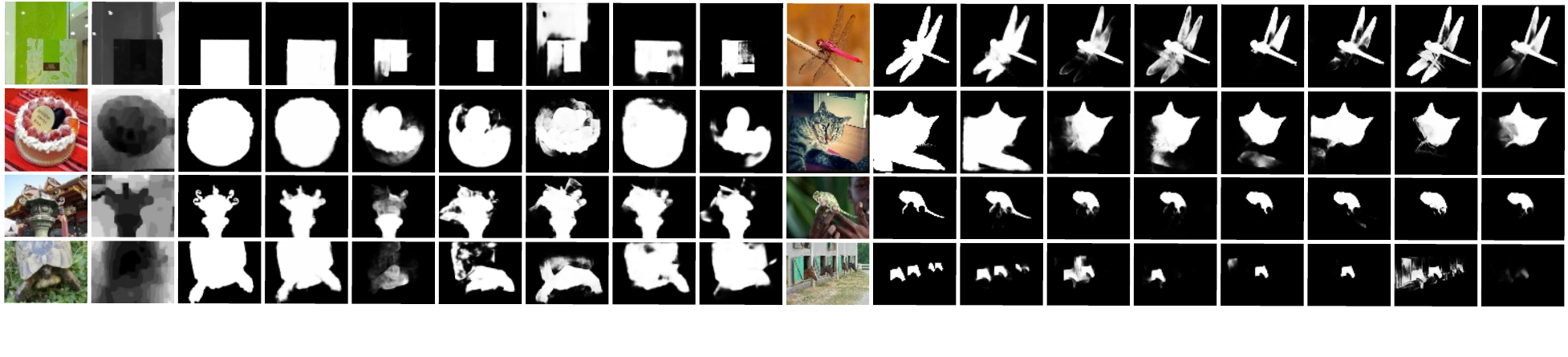}
  \put(1.2,1.6){\scriptsize Image}
  \put(6.9,1.6){\scriptsize Depth}
  \put(13.3,1.6){\scriptsize GT}
  \put(18.2,1.6){\scriptsize VST}
  \put(22.5,0){\scriptsize \shortstack[c] {BBS-Net\\ \cite{fan2020bbsnet}}}
  \put(28.8,0){\scriptsize \shortstack[c] {CoNet\\ \cite{Wei2020CoNet}}}
  \put(33.7,0){\scriptsize \shortstack[c] {HDFNet\\ \cite{HDFNet-ECCV2020}}}
  \put(39.7,0){\scriptsize \shortstack[c] {JLDCF\\ \cite{Fu2020JLDCF}}}
  \put(45,0){\scriptsize \shortstack[c] {UC-Net\\ \cite{zhang2020ucnet}}}
  \put(51.2, 1.6){\scriptsize Image}
  \put(57.5,1.6){\scriptsize GT}
  \put(62.5,1.6){\scriptsize VST}
  \put(67.2,0){\scriptsize \shortstack[c] {GateNet\\ \cite{GateNet}}}
  \put(73.7,0){\scriptsize \shortstack[c] {CSF\\ \cite{gao2020sod100k}}}
  \put(79.2,0){\scriptsize \shortstack[c] {LDF\\ \cite{CVPR2020_LDF}}}
  \put(84.2,0){\scriptsize \shortstack[c] {MINet\\ \cite{MINet-CVPR2020}}}
 \put(90.3,0){\scriptsize \shortstack[c] {ITSD\\ \cite{Zhou2020ITSD}}}
  \put(95.2,0){\scriptsize \shortstack[c] {EGNet\\ \cite{zhao2019EGNet}}}
  \end{overpic}
  \caption{Qualitative comparison against state-of-the-art RGB-D (left) and RGB (right) SOD methods. (GT: ground truth)}
  \label{visualcmp}
  \vspace{-0.3cm}
\end{figure*}
\vspace{-3mm}
\paragraph{Effectiveness of the multi-task transformer decoder.}
Based on ``+CMT+RT2T+F'', we further use our token-based multi-task decoder (TMD) to jointly perform saliency and boundary detection (``+CMT+RT2T+F+TMD"). It shows that using boundary detection can bring further performance gain for SOD on three out of four datasets. To very the effectiveness of our token-based prediction scheme, we try to directly use a conventional two-stream decoder (C2D) by using the ``+RT2T+F" architecture twice to predict the saliency map and boundary map via MLP, without using task-related tokens. This model is denoted as ``+CMT+RT2T+F+C2D'' in Table~\ref{ablationTab}.
The parameters and MACs of TMD vs. C2D are 17.22 M vs. 20.35 M and 17.70 G vs. 28.27 G, respectively.
The results show that using our TMD can achieve better results than using C2D on three out of four datasets, and also with much less computational costs. This clearly demonstrates the superiority of our proposed token-based transformer decoder.

\subsection{Comparison with State-of-the-Art Methods}
For RGB-D SOD, we compare our VST with 14 state-of-the-art RGB-D SOD methods, \ie, A2dele \cite{piao2020a2dele}, JL-DCF \cite{Fu2020JLDCF}, SSF-RGBD \cite{zhang2020select}, UC-Net \cite{zhang2020ucnet}, $S^2$MA \cite{liu2020S2MA}, PGAR \cite{chen2020PGAR}, DANet \cite{zhao2020DANet}, cmMS \cite{li2020cmMS}, ATSA \cite{zhang2020ATSA}, CMW \cite{Li2020CMWNet}, Cas-Gnn \cite{luo2020Cas-Gnn}, HDFNet \cite{HDFNet-ECCV2020}, CoNet \cite{Wei2020CoNet}, and BBS-Net \cite{fan2020bbsnet}.
For RGB SOD, we compare our VST with 12 state-of-the-art RGB SOD models, including GateNet \cite{GateNet}, CSF \cite{gao2020sod100k}, LDF \cite{CVPR2020_LDF}, MINet \cite{MINet-CVPR2020}, ITSD \cite{Zhou2020ITSD}, EGNet \cite{zhao2019EGNet}, TSPOANet \cite{Liu_TSPOANet}, AFNet \cite{Feng_AFNet}, PoolNet \cite{Liu19PoolNet}, CPD \cite{Wu_CPD}, BASNet \cite{Qin19BASNet}, and PiCANet \cite{liu2018picanet}. 
Table~\ref{RGBD_SOTA} and Table~\ref{RGB_SOTA} show the quantitative comparison results for RGB-D and RGB SOD, respectively.
The results show that our VST outperforms all previous state-of-the-art CNN-based SOD models on both RGB and RGB-D benchmark datasets, with comparable number of parameters and relatively small MACs, hence demonstrating the great effectiveness of our VST.
We also show visual comparison results among best-performed models in Figure~\ref{visualcmp}. It shows our proposed VST can accurately detect salient objects in very challenging scenarios, \eg, big salient objects, cluttered backgrounds, foreground and background having similar appearances, etc.

\section{Conclusion}
In this paper, we are the first to rethink SOD from a sequence-to-sequence perspective and develop a novel unified model based on a pure transformer, for both RGB and RGB-D SOD. To handle the difficulty of applying transformers in dense prediction tasks, we propose a new token upsampling method under the transformer framework and fuse multi-level patch tokens. We also design a multi-task decoder by introducing task-related tokens and a novel patch-task-attention mechanism to jointly perform saliency and boundary detection. Our VST model achieves state-of-the-art results for both RGB and RGB-D SOD without relying on heavy computational costs, thus showing its great effectiveness. We also set a new paradigm for the open question of how to use transformer in dense prediction tasks. 

\vspace{-4mm}
\paragraph{Acknowledgments:}
This work was supported in part by the National Key R\&D Program of China under Grant 2020AAA0105702, the National Science Foundation of China under Grant 62027813, 62036005, U20B2065, U20B2068.

{\small
\bibliographystyle{ieee_fullname}
\bibliography{egbib}

\begin{thebibliography}{10}\itemsep=-1pt

\bibitem{ba2016layer}
Jimmy~Lei Ba, Jamie~Ryan Kiros, and Geoffrey~E Hinton.
\newblock Layer normalization.
\newblock {\em arXiv preprint arXiv:1607.06450}, 2016.

\bibitem{borji2012exploiting}
Ali Borji and Laurent Itti.
\newblock Exploiting local and global patch rarities for saliency detection.
\newblock In {\em CVPR}, pages 478--485, 2012.

\bibitem{carion2020end}
Nicolas Carion, Francisco Massa, Gabriel Synnaeve, Nicolas Usunier, Alexander
  Kirillov, and Sergey Zagoruyko.
\newblock End-to-end object detection with transformers.
\newblock In {\em ECCV}, pages 213--229, 2020.

\bibitem{chen2018progressively}
Hao Chen and Youfu Li.
\newblock Progressively complementarity-aware fusion network for rgb-d salient
  object detection.
\newblock In {\em CVPR}, pages 3051--3060, 2018.

\bibitem{chen2019three}
Hao Chen and Youfu Li.
\newblock Three-stream attention-aware network for rgb-d salient object
  detection.
\newblock {\em TIP}, 28(6):2825--2835, 2019.

\bibitem{chen2020PGAR}
Shuhan Chen and Yun Fu.
\newblock Progressively guided alternate refinement network for rgb-d salient
  object detection.
\newblock In {\em ECCV}, pages 520--538, 2020.

\bibitem{chen2020dpanet}
Zuyao Chen, Runmin Cong, Qianqian Xu, and Qingming Huang.
\newblock Dpanet: Depth potentiality-aware gated attention network for rgb-d
  salient object detection.
\newblock {\em TIP}, 2020.

\bibitem{cheng2014global}
Ming-Ming Cheng, Niloy~J Mitra, Xiaolei Huang, Philip~HS Torr, and Shi-Min Hu.
\newblock Global contrast based salient region detection.
\newblock {\em TPAMI}, 37(3):569--582, 2014.

\bibitem{cheng2014rgbd135}
Yupeng Cheng, Huazhu Fu, Xingxing Wei, Jiangjian Xiao, and Xiaochun Cao.
\newblock Depth enhanced saliency detection method.
\newblock In {\em Conference on Internet Multimedia Computing and Service},
  pages 23--27, 2014.

\bibitem{2020performer}
Krzysztof Choromanski, Valerii Likhosherstov, David Dohan, Xingyou Song,
  Andreea Gane, Tamas Sarlos, Peter Hawkins, Jared Davis, Afroz Mohiuddin,
  Lukasz Kaiser, et~al.
\newblock Rethinking attention with performers.
\newblock In {\em ICLR}, 2020.

\bibitem{deng2018r3net}
Zijun Deng, Xiaowei Hu, Lei Zhu, Xuemiao Xu, Jing Qin, Guoqiang Han, and
  Pheng-Ann Heng.
\newblock R3net: Recurrent residual refinement network for saliency detection.
\newblock In {\em IJCAI}, pages 684--690, 2018.

\bibitem{dosovitskiy2020image}
Alexey Dosovitskiy, Lucas Beyer, Alexander Kolesnikov, Dirk Weissenborn,
  Xiaohua Zhai, Thomas Unterthiner, Mostafa Dehghani, Matthias Minderer, Georg
  Heigold, Sylvain Gelly, et~al.
\newblock An image is worth 16x16 words: Transformers for image recognition at
  scale.
\newblock In {\em ICLR}, 2020.

\bibitem{fan2017structure}
Deng-Ping Fan, Ming-Ming Cheng, Yun Liu, Tao Li, and Ali Borji.
\newblock Structure-measure: A new way to evaluate foreground maps.
\newblock In {\em ICCV}, pages 4548--4557, 2017.

\bibitem{Fan2018Enhanced}
Deng-Ping Fan, Cheng Gong, Yang Cao, Bo Ren, Ming-Ming Cheng, and Ali Borji.
\newblock {Enhanced-alignment Measure for Binary Foreground Map Evaluation}.
\newblock In {\em IJCAI}, pages 698--704, 2018.

\bibitem{fan2020SIP}
Deng-Ping Fan, Zheng Lin, Zhao Zhang, Menglong Zhu, and Ming-Ming Cheng.
\newblock Rethinking rgb-d salient object detection: Models, data sets, and
  large-scale benchmarks.
\newblock {\em TNNLS}, 32(5):2075--2089, 2020.

\bibitem{fan2020bbsnet}
Deng-Ping Fan, Yingjie Zhai, Ali Borji, Jufeng Yang, and Ling Shao.
\newblock Bbs-net: Rgb-d salient object detection with a bifurcated backbone
  strategy network.
\newblock In {\em ECCV}, pages 275--292, 2020.

\bibitem{Feng_AFNet}
Mengyang Feng, Huchuan Lu, and Errui Ding.
\newblock Attentive feedback network for boundary-aware salient object
  detection.
\newblock In {\em CVPR}, pages 1623--1632, 2019.

\bibitem{Fu2020JLDCF}
Keren Fu, Deng-Ping Fan, Ge-Peng Ji, and Qijun Zhao.
\newblock Jl-dcf: Joint learning and densely-cooperative fusion framework for
  rgb-d salient object detection.
\newblock In {\em CVPR}, pages 3052--3062, 2020.

\bibitem{gan2015devnet}
Chuang Gan, Naiyan Wang, Yi Yang, Dit-Yan Yeung, and Alex~G Hauptmann.
\newblock Devnet: A deep event network for multimedia event detection and
  evidence recounting.
\newblock In {\em CVPR}, pages 2568--2577, 2015.

\bibitem{gao2020sod100k}
Shang-Hua Gao, Yong-Qiang Tan, Ming-Ming Cheng, Chengze Lu, Yunpeng Chen, and
  Shuicheng Yan.
\newblock Highly efficient salient object detection with 100k parameters.
\newblock In {\em ECCV}, pages 702--721, 2020.

\bibitem{goferman2011context}
Stas Goferman, Lihi Zelnik-Manor, and Ayellet Tal.
\newblock Context-aware saliency detection.
\newblock {\em TPAMI}, 34(10):1915--1926, 2011.

\bibitem{han2017cnns}
Junwei Han, Hao Chen, Nian Liu, Chenggang Yan, and Xuelong Li.
\newblock Cnns-based rgb-d saliency detection via cross-view transfer and
  multiview fusion.
\newblock {\em IEEE transactions on cybernetics}, 48(11):3171--3183, 2017.

\bibitem{he2016resnet}
Kaiming He, Xiangyu Zhang, Shaoqing Ren, and Jian Sun.
\newblock Deep residual learning for image recognition.
\newblock In {\em CVPR}, pages 770--778, 2016.

\bibitem{hou2018dss}
Q Hou, MM Cheng, X Hu, A Borji, Z Tu, and PHS Torr.
\newblock Deeply supervised salient object detection with short connections.
\newblock {\em TPAMI}, 41(4):815--828, 2018.

\bibitem{Wei2020CoNet}
Wei Ji, Jingjing Li, Miao Zhang, Yongri Piao, and Huchuan Lu.
\newblock Accurate rgb-d salient object detection via collaborative learning.
\newblock In {\em ECCV}, pages 52--69, 2020.

\bibitem{ju2014njud}
Ran Ju, Ling Ge, Wenjing Geng, Tongwei Ren, and Gangshan Wu.
\newblock Depth saliency based on anisotropic center-surround difference.
\newblock In {\em ICIP}, pages 1115--1119, 2014.

\bibitem{Adam2015}
Diederik~P. Kingma and Jimmy Ba.
\newblock Adam: {A} method for stochastic optimization.
\newblock In {\em ICLR}, 2015.

\bibitem{lecun1998gradient}
Yann LeCun, L{\'e}on Bottou, Yoshua Bengio, and Patrick Haffner.
\newblock Gradient-based learning applied to document recognition.
\newblock {\em Proceedings of the IEEE}, 86(11):2278--2324, 1998.

\bibitem{li2020cmMS}
Chongyi Li, Runmin Cong, Yongri Piao, Qianqian Xu, and Chen~Change Loy.
\newblock Rgb-d salient object detection with cross-modality modulation and
  selection.
\newblock In {\em ECCV}, pages 225--241, 2020.

\bibitem{li2020icnet}
Gongyang Li, Zhi Liu, and Haibin Ling.
\newblock Icnet: Information conversion network for rgb-d based salient object
  detection.
\newblock {\em TIP}, 29:4873--4884, 2020.

\bibitem{Li2020CMWNet}
Gongyang Li, Zhi Liu, Linwei Ye, Yang Wang, and Haibin Ling.
\newblock Cross-modal weighting network for rgb-d salient object detection.
\newblock In {\em ECCV}, pages 665--681, 2020.

\bibitem{li2015HKUIS}
Guanbin Li and Yizhou Yu.
\newblock Visual saliency based on multiscale deep features.
\newblock In {\em CVPR}, pages 5455--5463, 2015.

\bibitem{li2014lfsd}
Nianyi Li, Jinwei Ye, Yu Ji, Haibin Ling, and Jingyi Yu.
\newblock Saliency detection on light field.
\newblock In {\em CVPR}, pages 2806--2813, 2014.

\bibitem{li2014PASCALS}
Yin Li, Xiaodi Hou, Christof Koch, James~M Rehg, and Alan~L Yuille.
\newblock The secrets of salient object segmentation.
\newblock In {\em CVPR}, pages 280--287, 2014.

\bibitem{Liu19PoolNet}
Jiang-Jiang Liu, Qibin Hou, Ming-Ming Cheng, Jiashi Feng, and Jianmin Jiang.
\newblock A simple pooling-based design for real-time salient object detection.
\newblock In {\em CVPR}, pages 3917--3926, 2019.

\bibitem{liu2016dhsnet}
Nian Liu and Junwei Han.
\newblock Dhsnet: Deep hierarchical saliency network for salient object
  detection.
\newblock In {\em CVPR}, pages 678--686, 2016.

\bibitem{liu2018picanet}
Nian Liu, Junwei Han, and Ming-Hsuan Yang.
\newblock Picanet: Learning pixel-wise contextual attention for saliency
  detection.
\newblock In {\em CVPR}, pages 3089--3098, 2018.

\bibitem{liu2020S2MA}
Nian Liu, Ni Zhang, and Junwei Han.
\newblock Learning selective self-mutual attention for rgb-d saliency
  detection.
\newblock In {\em CVPR}, pages 13756--13765, 2020.

\bibitem{liu2020ReDWeb-S}
Nian Liu, Ni Zhang, Ling Shao, and Junwei Han.
\newblock Learning selective mutual attention and contrast for rgb-d saliency
  detection.
\newblock {\em arXiv preprint arXiv:2010.05537}, 2020.

\bibitem{liu2021end}
Ruijin Liu, Zejian Yuan, Tie Liu, and Zhiliang Xiong.
\newblock End-to-end lane shape prediction with transformers.
\newblock In {\em IEEE Winter Conference on Applications of Computer Vision},
  pages 3694--3702, 2021.

\bibitem{Liu_TSPOANet}
Yi Liu, Qiang Zhang, Dingwen Zhang, and Jungong Han.
\newblock Employing deep part-object relationships for salient object
  detection.
\newblock In {\em ICCV}, pages 1232--1241, 2019.

\bibitem{liu2019salient}
Zhengyi Liu, Song Shi, Quntao Duan, Wei Zhang, and Peng Zhao.
\newblock Salient object detection for rgb-d image by single stream recurrent
  convolution neural network.
\newblock {\em Neurocomputing}, 363:46--57, 2019.

\bibitem{luo2020Cas-Gnn}
Ao Luo, Xin Li, Fan Yang, Zhicheng Jiao, Hong Cheng, and Siwei Lyu.
\newblock Cascade graph neural networks for rgb-d salient object detection.
\newblock In {\em ECCV}, pages 346--364, 2020.

\bibitem{luo2017non}
Zhiming Luo, Akshaya Mishra, Andrew Achkar, Justin Eichel, Shaozi Li, and
  Pierre-Marc Jodoin.
\newblock Non-local deep features for salient object detection.
\newblock In {\em CVPR}, pages 6609--6617, 2017.

\bibitem{movahedi2010SOD}
Vida Movahedi and James~H Elder.
\newblock Design and perceptual validation of performance measures for salient
  object segmentation.
\newblock In {\em CVPR Workshops}, pages 49--56, 2010.

\bibitem{niu2012stere}
Yuzhen Niu, Yujie Geng, Xueqing Li, and Feng Liu.
\newblock Leveraging stereopsis for saliency analysis.
\newblock In {\em CVPR}, pages 454--461, 2012.

\bibitem{noh2015learning}
Hyeonwoo Noh, Seunghoon Hong, and Bohyung Han.
\newblock Learning deconvolution network for semantic segmentation.
\newblock In {\em ICCV}, pages 1520--1528, 2015.

\bibitem{HDFNet-ECCV2020}
Youwei Pang, Lihe Zhang, Xiaoqi Zhao, and Huchuan Lu.
\newblock Hierarchical dynamic filtering network for rgb-d salient object
  detection.
\newblock In {\em ECCV}, pages 235--252, 2020.

\bibitem{MINet-CVPR2020}
Youwei Pang, Xiaoqi Zhao, Lihe Zhang, and Huchuan Lu.
\newblock Multi-scale interactive network for salient object detection.
\newblock In {\em CVPR}, pages 9413--9422, 2020.

\bibitem{paszke2019pytorch}
Adam Paszke, Sam Gross, Francisco Massa, Adam Lerer, James Bradbury, Gregory
  Chanan, Trevor Killeen, Zeming Lin, Natalia Gimelshein, Luca Antiga, et~al.
\newblock Pytorch: An imperative style, high-performance deep learning library.
\newblock {\em NIPS}, 32:8026--8037, 2019.

\bibitem{peng2014nlpr}
Houwen Peng, Bing Li, Weihua Xiong, Weiming Hu, and Rongrong Ji.
\newblock Rgbd salient object detection: A benchmark and algorithms.
\newblock In {\em ECCV}, pages 92--109, 2014.

\bibitem{Piao2019dmra}
Yongri Piao, Wei Ji, Jingjing Li, Miao Zhang, and Huchuan Lu.
\newblock Depth-induced multi-scale recurrent attention network for saliency
  detection.
\newblock In {\em ICCV}, pages 7254--7263, 2019.

\bibitem{piao2020a2dele}
Yongri Piao, Zhengkun Rong, Miao Zhang, Weisong Ren, and Huchuan Lu.
\newblock A2dele: Adaptive and attentive depth distiller for efficient rgb-d
  salient object detection.
\newblock In {\em CVPR}, pages 9060--9069, 2020.

\bibitem{qin2019basnet}
Xuebin Qin, Zichen Zhang, Chenyang Huang, Chao Gao, Masood Dehghan, and Martin
  Jagersand.
\newblock Basnet: Boundary-aware salient object detection.
\newblock In {\em CVPR}, pages 7479--7489, 2019.

\bibitem{Qin19BASNet}
Xuebin Qin, Zichen Zhang, Chenyang Huang, Chao Gao, Masood Dehghan, and Martin
  Jagersand.
\newblock Basnet: Boundary-aware salient object detection.
\newblock In {\em CVPR}, pages 7479--7489, 2019.

\bibitem{ren2015exploiting}
Jianqiang Ren, Xiaojin Gong, Lu Yu, Wenhui Zhou, and Michael Ying~Yang.
\newblock Exploiting global priors for rgb-d saliency detection.
\newblock In {\em CVPR workshops}, pages 25--32, 2015.

\bibitem{ronneberger2015unet}
Olaf Ronneberger, Philipp Fischer, and Thomas Brox.
\newblock U-net: Convolutional networks for biomedical image segmentation.
\newblock In {\em International Conference on Medical image computing and
  computer-assisted intervention}, pages 234--241, 2015.

\bibitem{shimoda2016distinct}
Wataru Shimoda and Keiji Yanai.
\newblock Distinct class-specific saliency maps for weakly supervised semantic
  segmentation.
\newblock In {\em ECCV}, pages 218--234, 2016.

\bibitem{simonyan2014vgg}
Karen Simonyan and Andrew Zisserman.
\newblock Very deep convolutional networks for large-scale image recognition.
\newblock In {\em ICLR}, 2015.

\bibitem{touvron2020training}
Hugo Touvron, Matthieu Cord, Matthijs Douze, Francisco Massa, Alexandre
  Sablayrolles, and Herv{\'e} J{\'e}gou.
\newblock Training data-efficient image transformers \& distillation through
  attention.
\newblock In {\em ICML}, pages 10347--10357, 2021.

\bibitem{vaswani2017attention}
Ashish Vaswani, Noam Shazeer, Niki Parmar, Jakob Uszkoreit, Llion Jones,
  Aidan~N Gomez, {\L}ukasz Kaiser, and Illia Polosukhin.
\newblock Attention is all you need.
\newblock In {\em NIPS}, pages 5998--6008, 2017.

\bibitem{wang2020maxdeeplab}
Huiyu Wang, Yukun Zhu, Hartwig Adam, Alan Yuille, and Liang-Chieh Chen.
\newblock Max-deeplab: End-to-end panoptic segmentation with mask transformers.
\newblock In {\em CVPR}, pages 5463--5474, 2021.

\bibitem{wang2017duts}
Lijun Wang, Huchuan Lu, Yifan Wang, Mengyang Feng, Dong Wang, Baocai Yin, and
  Xiang Ruan.
\newblock Learning to detect salient objects with image-level supervision.
\newblock In {\em CVPR}, pages 136--145, 2017.

\bibitem{wang2018rfcn}
Linzhao Wang, Lijun Wang, Huchuan Lu, Pingping Zhang, and Xiang Ruan.
\newblock Salient object detection with recurrent fully convolutional networks.
\newblock {\em TPAMI}, 41(7):1734--1746, 2018.

\bibitem{wang2017stagewise}
Tiantian Wang, Ali Borji, Lihe Zhang, Pingping Zhang, and Huchuan Lu.
\newblock A stagewise refinement model for detecting salient objects in images.
\newblock In {\em ICCV}, pages 4019--4028, 2017.

\bibitem{wang2019salient1}
Wenguan Wang, Qiuxia Lai, Huazhu Fu, Jianbing Shen, and Haibin Ling.
\newblock {Salient object detection in the deep learning era: An in-depth
  survey}.
\newblock {\em TPAMI}, 2021.

\bibitem{wang2018salient}
Wenguan Wang, Jianbing Shen, Xingping Dong, and Ali Borji.
\newblock Salient object detection driven by fixation prediction.
\newblock In {\em CVPR}, pages 1711--1720, 2018.

\bibitem{wang2021pvt}
Wenhai Wang, Enze Xie, Xiang Li, Deng-Ping Fan, Kaitao Song, Ding Liang, Tong
  Lu, Ping Luo, and Ling Shao.
\newblock Pyramid vision transformer: A versatile backbone for dense prediction
  without convolutions.
\newblock {\em arXiv preprint arXiv:2102.12122}, 2021.

\bibitem{CVPR2020_LDF}
Jun Wei, Shuhui Wang, Zhe Wu, Chi Su, Qingming Huang, and Qi Tian.
\newblock Label decoupling framework for salient object detection.
\newblock In {\em CVPR}, pages 13025--13034, 2020.

\bibitem{Wu_CPD}
Zhe Wu, Li Su, and Qingming Huang.
\newblock Cascaded partial decoder for fast and accurate salient object
  detection.
\newblock In {\em CVPR}, pages 3907--3916, 2019.

\bibitem{xie2015hed}
Saining Xie and Zhuowen Tu.
\newblock Holistically-nested edge detection.
\newblock In {\em CVPR}, pages 1395--1403, 2015.

\bibitem{yan2013ECSSD}
Qiong Yan, Li Xu, Jianping Shi, and Jiaya Jia.
\newblock Hierarchical saliency detection.
\newblock In {\em CVPR}, pages 1155--1162, 2013.

\bibitem{yang2013DUTO}
Chuan Yang, Lihe Zhang, Huchuan Lu, Xiang Ruan, and Ming-Hsuan Yang.
\newblock Saliency detection via graph-based manifold ranking.
\newblock In {\em CVPR}, pages 3166--3173, 2013.

\bibitem{yuan2021tokens}
Li Yuan, Yunpeng Chen, Tao Wang, Weihao Yu, Yujun Shi, Francis~EH Tay, Jiashi
  Feng, and Shuicheng Yan.
\newblock Tokens-to-token vit: Training vision transformers from scratch on
  imagenet.
\newblock In {\em ICCV}, 2021.

\bibitem{zhai2006visual}
Yun Zhai and Mubarak Shah.
\newblock Visual attention detection in video sequences using spatiotemporal
  cues.
\newblock In {\em ACM International Conference on Multimedia}, pages 815--824,
  2006.

\bibitem{zhang2020ucnet}
Jing Zhang, Deng-Ping Fan, Yuchao Dai, Saeed Anwar, Fatemeh~Sadat Saleh, Tong
  Zhang, and Nick Barnes.
\newblock Uc-net: Uncertainty inspired rgb-d saliency detection via conditional
  variational autoencoders.
\newblock In {\em CVPR}, pages 8582--8591, 2020.

\bibitem{zhang2019capsal}
Lu Zhang, Jianming Zhang, Zhe Lin, Huchuan Lu, and You He.
\newblock Capsal: Leveraging captioning to boost semantics for salient object
  detection.
\newblock In {\em CVPR}, pages 6024--6033, 2019.

\bibitem{zhang2020ATSA}
Miao Zhang, Sun~Xiao Fei, Jie Liu, Shuang Xu, Yongri Piao, and Huchuan Lu.
\newblock Asymmetric two-stream architecture for accurate rgb-d saliency
  detection.
\newblock In {\em ECCV}, pages 374--390, 2020.

\bibitem{zhang2020select}
Miao Zhang, Weisong Ren, Yongri Piao, Zhengkun Rong, and Huchuan Lu.
\newblock Select, supplement and focus for rgb-d saliency detection.
\newblock In {\em CVPR}, pages 3472--3481, 2020.

\bibitem{zhang2018pagr}
Xiaoning Zhang, Tiantian Wang, Jinqing Qi, Huchuan Lu, and Gang Wang.
\newblock Progressive attention guided recurrent network for salient object
  detection.
\newblock In {\em CVPR}, pages 714--722, 2018.

\bibitem{zhao2019contrast}
Jia-Xing Zhao, Yang Cao, Deng-Ping Fan, Ming-Ming Cheng, Xuan-Yi Li, and Le
  Zhang.
\newblock Contrast prior and fluid pyramid integration for rgbd salient object
  detection.
\newblock In {\em CVPR}, pages 3927--3936, 2019.

\bibitem{zhao2019EGNet}
Jia-Xing Zhao, Jiang-Jiang Liu, Deng-Ping Fan, Yang Cao, Jufeng Yang, and
  Ming-Ming Cheng.
\newblock Egnet:edge guidance network for salient object detection.
\newblock In {\em ICCV}, pages 8779--8788, 2019.

\bibitem{zhao2015saliency}
Rui Zhao, Wanli Ouyang, Hongsheng Li, and Xiaogang Wang.
\newblock Saliency detection by multi-context deep learning.
\newblock In {\em CVPR}, pages 1265--1274, 2015.

\bibitem{GateNet}
Xiaoqi Zhao, Youwei Pang, Lihe Zhang, Huchuan Lu, and Lei Zhang.
\newblock Suppress and balance: A simple gated network for salient object
  detection.
\newblock In {\em ECCV}, pages 35--51, 2020.

\bibitem{zhao2020DANet}
Xiaoqi Zhao, Lihe Zhang, Youwei Pang, Huchuan Lu, and Lei Zhang.
\newblock A single stream network for robust and real-time rgb-d salient object
  detection.
\newblock In {\em ECCV}, pages 646--662, 2020.

\bibitem{zheng2020rethinking}
Sixiao Zheng, Jiachen Lu, Hengshuang Zhao, Xiatian Zhu, Zekun Luo, Yabiao Wang,
  Yanwei Fu, Jianfeng Feng, Tao Xiang, Philip~HS Torr, et~al.
\newblock Rethinking semantic segmentation from a sequence-to-sequence
  perspective with transformers.
\newblock In {\em CVPR}, pages 6881--6890, 2021.

\bibitem{Zhou2020ITSD}
Huajun Zhou, Xiaohua Xie, Jian-Huang Lai, Zixuan Chen, and Lingxiao Yang.
\newblock Interactive two-stream decoder for accurate and fast saliency
  detection.
\newblock In {\em CVPR}, pages 9141--9150, 2020.

\bibitem{zhou2021rgb}
Tao Zhou, Deng-Ping Fan, Ming-Ming Cheng, Jianbing Shen, and Ling Shao.
\newblock Rgb-d salient object detection: A survey.
\newblock {\em Computational Visual Media}, pages 1--33, 2021.

\bibitem{specificity_rgbd_sod}
Tao Zhou, Huazhu Fu, Geng Chen, Yi Zhou, Deng-Ping Fan, and Ling Shao.
\newblock Specificity-preserving rgb-d saliency detection.
\newblock In {\em ICCV}, 2021.

\bibitem{zhu2017ssd}
Chunbiao Zhu and Ge Li.
\newblock A three-pathway psychobiological framework of salient object
  detection using stereoscopic technology.
\newblock In {\em ICCV Workshops}, pages 3008--3014, 2017.

\bibitem{zhu2020deformable}
Xizhou Zhu, Weijie Su, Lewei Lu, Bin Li, Xiaogang Wang, and Jifeng Dai.
\newblock Deformable detr: Deformable transformers for end-to-end object
  detection.
\newblock In {\em ICLR}, 2020.

\end{thebibliography}
}

\clearpage

\begin{table*}[!ht]
\centering
\footnotesize
\renewcommand{\arraystretch}{1.1}
\renewcommand{\tabcolsep}{1.3mm}
\caption{Ablation studies of our proposed model on RGB SOD datasets. ``RC'' means RGB Convertor.  ``Bili'' denotes bilinear upsampling and ``F" means multi-level token fusion. ``TMD" denotes our proposed token-based multi-task decoder, while ``C2D'' means using conventional two-stream decoder to perform saliency and boundary detection without using task-related tokens. The best results are labeled in \blu{blue}.
}
\begin{tabular}{l|l|cccc|cccc|cccc|cccc}
\hline
\multicolumn{2}{l|}{\multirow{2}{*}{Settings}} & \multicolumn{4}{c|}{DUTS \cite{wang2017duts}} & \multicolumn{4}{c|}{HKU-IS \cite{li2015HKUIS}} & \multicolumn{4}{c|}{PASCAL-S \cite{li2014PASCALS}} & \multicolumn{4}{c}{SOD \cite{movahedi2010SOD}}\\
\multicolumn{2}{l|}{} & \multicolumn{1}{l}{$S_m$} & \multicolumn{1}{l}{maxF} & \multicolumn{1}{l}{$E_{\xi}^{\text{max}}$} & \multicolumn{1}{l|}{MAE}
                      & \multicolumn{1}{l}{$S_m$} & \multicolumn{1}{l}{maxF} & \multicolumn{1}{l}{$E_{\xi}^{\text{max}}$} & \multicolumn{1}{l|}{MAE}
                      & \multicolumn{1}{l}{$S_m$} & \multicolumn{1}{l}{maxF} & \multicolumn{1}{l}{$E_{\xi}^{\text{max}}$} & \multicolumn{1}{l|}{MAE}
                      & \multicolumn{1}{l}{$S_m$} & \multicolumn{1}{l}{maxF} & \multicolumn{1}{l}{$E_{\xi}^{\text{max}}$} & \multicolumn{1}{l}{MAE}

  \\ \hline

\multicolumn{2}{l|}{Baseline}  &0.824  &0.780  &0.909  &0.071 &0.858  &0.854  &0.938  &0.075 &0.826 &0.795 &0.878  &0.096 &0.802 &0.803  &0.880 &0.100 \\ \hline
\multicolumn{2}{l|}{+RC}      &0.827  &0.785  &0.913  &0.070 &0.860  &0.856  &0.939  &0.074 &0.830 &0.797 &0.879  &0.095 &0.804 &0.805  &0.880 &0.100  \\ \hline
\multicolumn{2}{l|}{+RC+Bili} &0.867  &0.835  &0.929  &0.048 &0.901  &0.901  &0.956  &0.044 &0.856 &0.827 &0.891  &0.074 &0.833 &0.836  &0.891 &0.077   \\
\multicolumn{2}{l|}{+RC+RT2T} &0.881  &0.856  &0.934  &0.043 &0.914  &0.918  &0.961  &0.037 &0.864 &0.838 &0.896  &0.070 &0.844 &0.850  &0.894 &0.069    \\ \hline
\multicolumn{2}{l|}{+RC+RT2T+F}  &0.895  &0.874  &\blu{0.939} &0.039 &0.925 &0.932 &0.966  &0.032 &0.871 &0.845 &0.897  &0.068 &0.851 &0.861  &0.899 &0.068  \\
\multicolumn{2}{l|}{+RC+RT2T+F+TMD}  &\blu{0.896}  &\blu{0.877} &\blu{0.939} &\blu{0.037} &\blu{0.928} &\blu{0.937} &\blu{0.968} &\blu{0.030} &\blu{0.873} &\blu{0.850} &\blu{0.900} &\blu{0.067} &\blu{0.854} &\blu{0.866} &\blu{0.902} &\blu{0.065}\\ \hline
\multicolumn{2}{l|}{+RC+RT2T+F+C2D}  &0.891  &0.870 &0.937 &0.040 &0.924 &0.931 &0.966 &0.033 &0.869 &0.844 &0.896 &0.069 &0.852 &0.860 &0.898 &0.067\\ \hline
\end{tabular}
\label{ablationTab2}
\end{table*}

\begin{table*}[t!]
\centering
\scriptsize
\renewcommand{\arraystretch}{1.1}
\renewcommand{\tabcolsep}{1.1mm}
\caption{Comparison of using different numbers of transformer layers in our VST model. The final model setting is labeled in \blu{blue}.}
\begin{tabular}{c|cccc|cr|cccc|cccc|cccc|cccc}
\hline
 ID & \multicolumn{4}{c|}{Layer Num}  & \multicolumn{1}{c}{MACs} & \multicolumn{1}{c|}{Params}
 & \multicolumn{4}{c|}{NJUD \cite{ju2014njud}} & \multicolumn{4}{c|}{DUTLF-Depth \cite{Piao2019dmra}} & \multicolumn{4}{c|}{STERE \cite{niu2012stere}}   & \multicolumn{4}{c}{LFSD \cite{li2014lfsd}}      \\
 & \multicolumn{1}{l}{$L^{\mathcal{C}}$} & \multicolumn{1}{l}{$L_3^{\mathcal{D}}$} & \multicolumn{1}{l}{$L_2^{\mathcal{D}}$} & \multicolumn{1}{l|}{$L_1^{\mathcal{D}}$}
& \multicolumn{1}{r}{(G)} & \multicolumn{1}{r|}{(M)}
& \multicolumn{1}{l}{$S_m$} & \multicolumn{1}{l}{maxF} & \multicolumn{1}{l}{$E_{\xi}^{\text{max}}$} & \multicolumn{1}{l|}{MAE}
& \multicolumn{1}{l}{$S_m$} & \multicolumn{1}{l}{maxF} & \multicolumn{1}{l}{$E_{\xi}^{\text{max}}$} & \multicolumn{1}{l|}{MAE}
& \multicolumn{1}{l}{$S_m$} & \multicolumn{1}{l}{maxF} & \multicolumn{1}{l}{$E_{\xi}^{\text{max}}$} & \multicolumn{1}{l|}{MAE}
& \multicolumn{1}{l}{$S_m$} & \multicolumn{1}{l}{maxF} & \multicolumn{1}{l}{$E_{\xi}^{\text{max}}$} & \multicolumn{1}{l}{MAE}
\\ \hline

I& 8 &8 &4 &4  &48.35 &119.30 &0.925 &0.925 &0.955 &0.033 &0.940  &0.947 &0.966 &0.026 	&0.910 &0.902 &0.948 &0.039 &0.878 &0.884 &0.914 &0.066 \\ \hline
II& 8 &8 &2 &2  &36.78 &113.39 &0.923 &0.922 &0.955 &0.035 &0.943  &0.947 &0.968 &0.025 	&0.911 &0.904 &0.948 &0.039 &0.874 &0.878 &0.908 &0.069  \\
III& 8 &6 &2 &2  &36.20 &110.43 &0.921 &0.920 &0.952 &0.036 &0.940  &0.945 &0.966 &0.026 	&0.910 &0.904 &0.948 &0.040 &0.875 &0.883 &0.911 &0.067  \\
IV& 8 &4 &2 &2  &35.61 &107.47 &0.921 &0.920 &0.951 &0.036 &0.942  &0.947 &0.968 &0.026 	&0.911 &0.904 &0.949 &0.040 &0.876 &0.880 &0.912 &0.068   \\
V& 8 &2 &2 &2  &35.03 &104.52 &0.922 &0.921 &0.952 &0.036 &0.940  &0.944 &0.965 &0.026 	&0.912 &0.906 &0.949 &0.039 &0.873 &0.875 &0.908 &0.068 \\ \hline
VI& 6 &4 &2 &2  &33.30 &95.65  &0.923 &0.921 &0.952 &0.036 &0.943  &0.948 &0.968 &0.024 	&0.913 &0.906 &0.949 &0.039 &0.875 &0.878 &0.912 &0.067\\
VII& \blu{4} &\blu{4} &\blu{2} &\blu{2}  &30.99 &83.83  &0.922 &0.920 &0.951 &0.035 &0.943  &0.948 &0.969 &0.024 	&0.913 &0.907 &0.951 &0.038 &0.882 &0.889 &0.921 &0.061\\
VIII& 2 &4 &2 &2  &28.68 &72.00  &0.923 &0.921 &0.953 &0.036 &0.938  &0.943 &0.963 &0.028 	&0.912 &0.906 &0.950 &0.039 &0.881 &0.887 &0.917 &0.062\\ \hline
\end{tabular}
\label{layerTab}
\vspace{-3mm}
\end{table*}

\section{Supplementary materials}
\subsection{Ablation Study on RGB SOD Datasets}

We further report the results of ablation studies on four RGB SOD datasets, \ie, DUTS, HKU-IS, PASCAL-S, and SOD, in Table~\ref{ablationTab2} to demonstrate the effectiveness of our VST model components.

The baseline model is using transformer encoder to extract patch tokens $\bm{T}_r^{\mathcal{E}}$ and then directly using $\bm{T}_r^{\mathcal{E}}$ to predict the saliency map with 1/16 scale by using MLP on each patch token.
Based on the baseline, we insert RGB convertor right after the transformer encoder, shown as ``+RC'' in Table~\ref{ablationTab2}.
Compared to the baseline, RC brings performance gains especially on the DUTS and PASCAL-S datasets, which demonstrates its effectiveness.
For other components, \ie, RT2T, multi-level token fusion, and multi-task transformer decoder, we get consistent conclusions with the ablation studies on RGB-D SOD datasets as follows.

First, using bilinear upsampling (``+RC+Bili") can significantly improve the model performance while using our proposed RT2T (``+RC+RT2T") can further bring performance gains, hence demonstrating the effectiveness of our proposed RT2T.
Second, based on ``+RC+RT2T'', multi-level token fusion (``+RC+RT2T+F") can lead to better performance on all four datasets, which verifies its effectiveness.
Third, using multi-task transformer decoder (``+RC+RT2T+F+TMD") can improve the model performance on all four datasets and it is also superior to the conventional two-stream decoder (``+RC+RT2T+F+C2D").

To this end, the results of ablation studies on both RGB and RGB-D SOD datasets strongly demonstrate the effectiveness of our proposed VST components.

\subsection{Layer Number Study}
We conduct experiments to study the optimal numbers of different transformer layers, \ie, $L^{\mathcal{C}}$ in the transformer convertor and $L^{\mathcal{D}}$ in the multi-task transformer decoder, jointly considering computational costs and model performance. Note that there are three decoder modules at three scales in the multi-task transformer decoder, thus we set different transformer layer numbers for them, \ie, $L_3^{\mathcal{D}}$ for 1/16 scale, $L_2^{\mathcal{D}}$ for 1/8 scale, and $L_1^{\mathcal{D}}$ for 1/4 scale.
The experimental results on four RGB-D SOD datasets, \ie, NJUD, DUTLF-Depth, STERE, and LFSD, are given in Table~\ref{layerTab}.

In our initial model setting, we set $L^{\mathcal{C}}=L_3^{\mathcal{D}}=8$. Since $L_2^{\mathcal{D}}$ and $L_1^{\mathcal{D}}$ are used at relatively large scales, we initially set both of them to 4, as shown in row I in Table~\ref{layerTab}. Then, we start to change the numbers of different layers.

We first reduce $L_2^{\mathcal{D}}$ and $L_1^{\mathcal{D}}$ from 4 to 2 to save computational costs.
The experimental results on row II show that it can get comparable performance with less computational costs compared with row I.
Hence, we set $L_2^{\mathcal{D}}=L_1^{\mathcal{D}}=2$ and start to change $L_3^{\mathcal{D}}$ from 8 to 6, 4, 2, respectively, which are shown in row III, IV, V in Table~\ref{layerTab}.
We find that as $L_3^{\mathcal{D}}$ decreases, the computation costs decrease gradually while the results are generally comparable.
However, the model performance on row IV is better than that on row V on DUTLF-Depth and LFSD datasets.
Thus, we set $L_3^{\mathcal{D}}=4$ and start to change $L^{\mathcal{C}}$ from 8 to 6, 4, 2, respectively, which are shown in row VI, VII, VIII.
It can be seen that the performance on row VII is the best and the model has acceptable computational costs.
Hence, we set $L^{\mathcal{C}}=L_3^{\mathcal{D}}=4$ and $L_2^{\mathcal{D}}=L_1^{\mathcal{D}}=2$ as our final model setting.

\subsection{More Visual Comparison with State-of-the-art Methods}
\begin{figure*}[t]
  \graphicspath{{Figures/qualitative/}}
  \centering
  \begin{overpic}[width=1\linewidth]{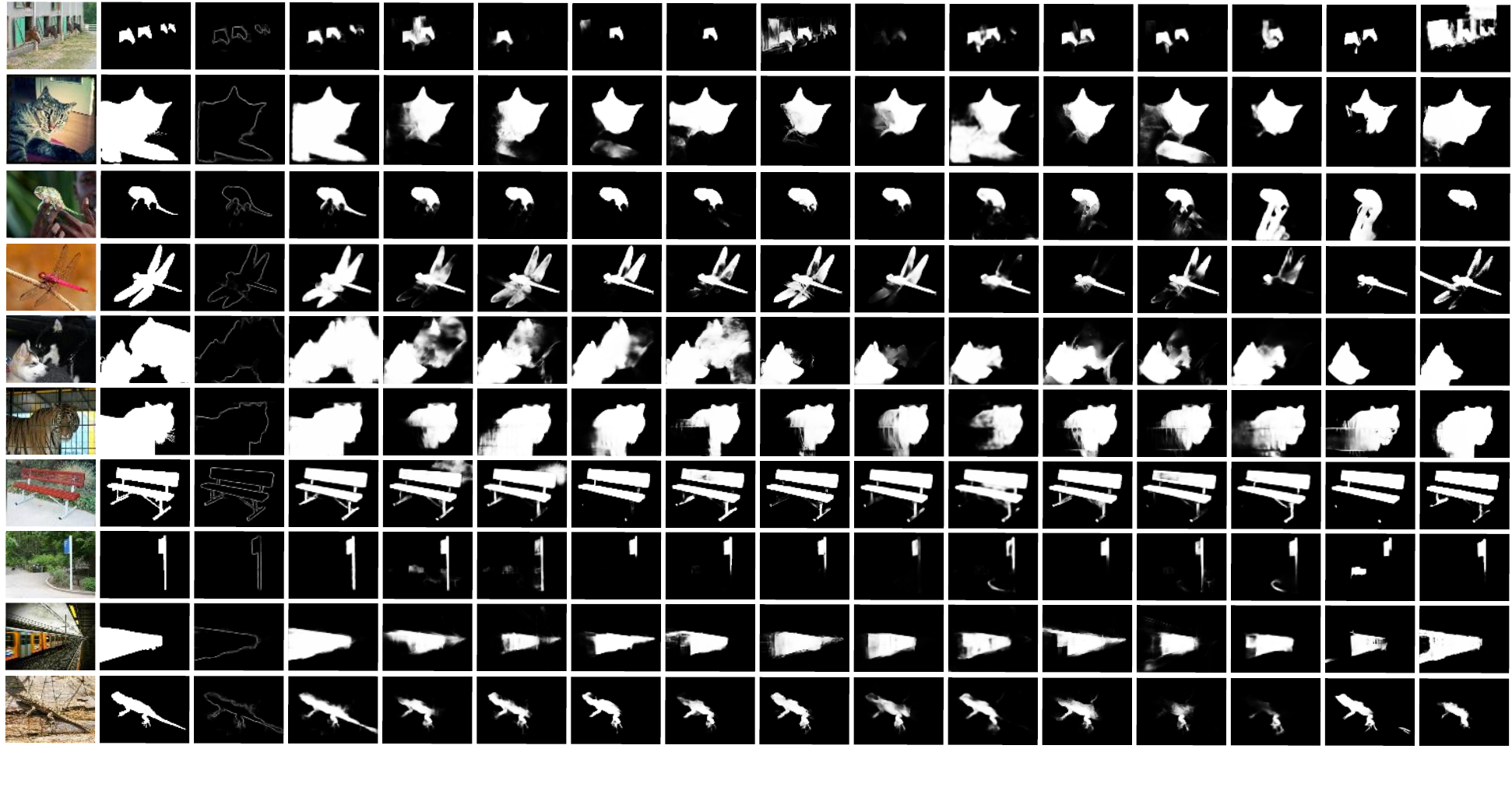}
  \put(1.5, 1.6){\scriptsize Image}
  \put(9.5,1.6){\scriptsize GT}
  \put(15.5,1.6){\scriptsize VST}
  \put(21,0){\scriptsize \shortstack[c] {GateNet\\ \cite{GateNet}}}
  \put(29,0){\scriptsize \shortstack[c] {CSF\\ \cite{gao2020sod100k}}}
  \put(35.1,0){\scriptsize \shortstack[c] {LDF\\ \cite{CVPR2020_LDF}}}
  \put(41.6,0){\scriptsize \shortstack[c] {MINet\\ \cite{MINet-CVPR2020}}}
  \put(48.5,0){\scriptsize \shortstack[c] {ITSD\\ \cite{Zhou2020ITSD}}}
  \put(54.8,0){\scriptsize \shortstack[c] {EGNet\\ \cite{zhao2019EGNet}}}
  \put(61.5,0){\scriptsize \shortstack[c] {TSPOA\\ \cite{Liu_TSPOANet}}}
  \put(68.2,0){\scriptsize \shortstack[c] {AFNet\\ \cite{Feng_AFNet}}}
  \put(74.8,0){\scriptsize \shortstack[c] {PoolNet\\ \cite{Liu19PoolNet}}}
  \put(82.3,0){\scriptsize \shortstack[c] {CPD\\ \cite{Wu_CPD}}}
  \put(87.8,0){\scriptsize \shortstack[c] {BASNet\\ \cite{Qin19BASNet}}}
  \put(94.5,0){\scriptsize \shortstack[c] {PiCANet\\ \cite{liu2018picanet}}}
  \end{overpic}
  \caption{Qualitative comparison against state-of-the-art RGB SOD methods. (GT: ground truth)}
  \label{visualcmpRGB}
  \vspace{-0.3cm}
\end{figure*}

\begin{figure*}[t]
  \graphicspath{{Figures/qualitative/}}
  \centering
  \begin{overpic}[width=1\linewidth]{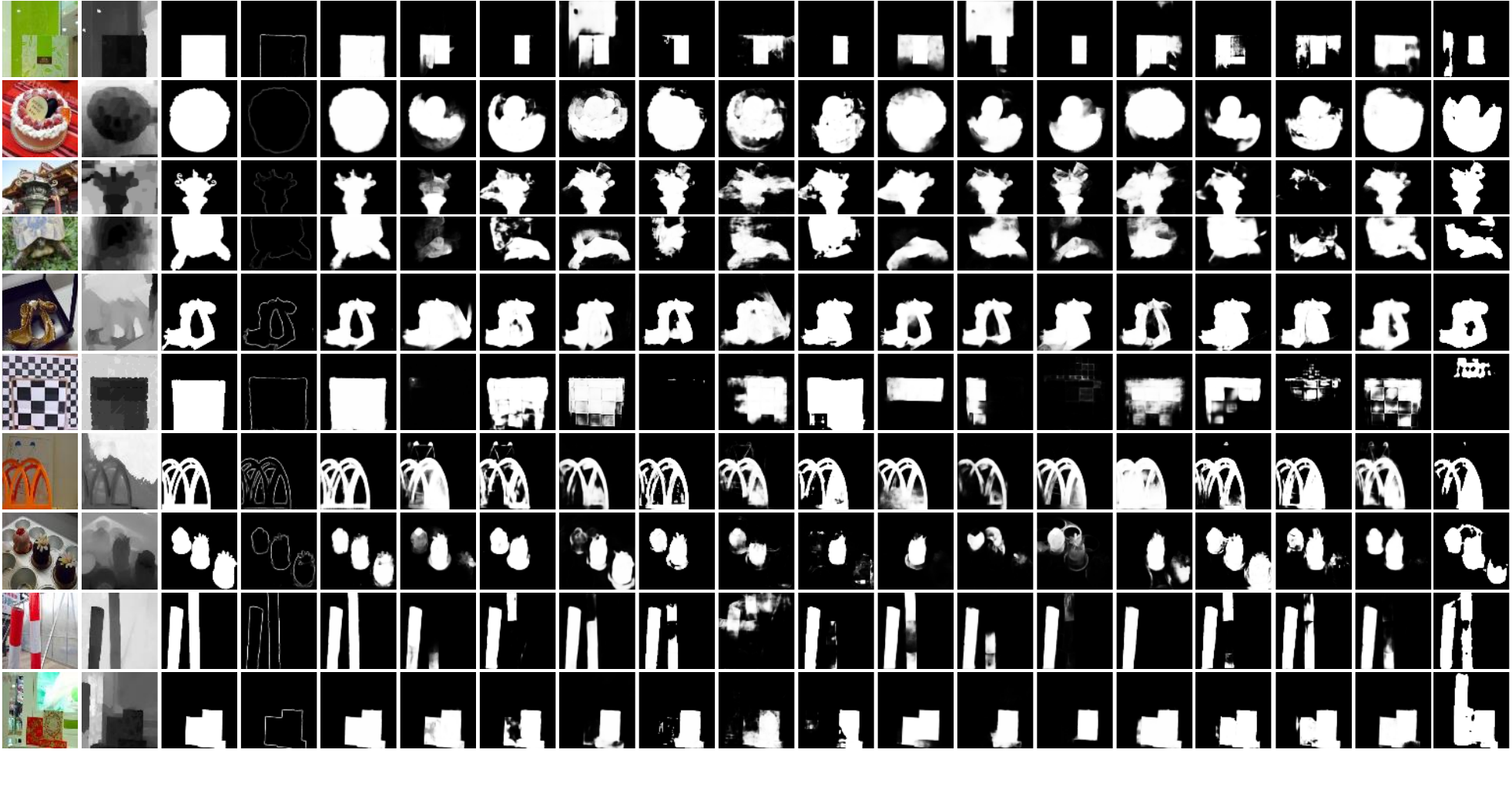}
  \put(1.3,1.6){\scriptsize Image}
  \put(7,1.6){\scriptsize Depth}
  \put(13.3,1.6){\scriptsize GT}
  \put(18.2,1.6){\scriptsize VST}
  \put(23,0){\scriptsize \shortstack[c] {BBS-Net\\ \cite{fan2020bbsnet}}}
  \put(29,0){\scriptsize \shortstack[c] {CoNet\\ \cite{Wei2020CoNet}}}
  \put(34,0){\scriptsize \shortstack[c] {HDFNet\\ \cite{HDFNet-ECCV2020}}}
  \put(39.5,0){\scriptsize \shortstack[c] {Cas-Gnn\\ \cite{luo2020Cas-Gnn}}}
  \put(45,0){\scriptsize \shortstack[c] {CMW\\ \cite{Li2020CMWNet}}}
  \put(50.8,0){\scriptsize \shortstack[c] {ATST\\ \cite{zhang2020ATSA}}}
  \put(56.5,0){\scriptsize \shortstack[c] {cmMS\\ \cite{li2020cmMS}}}
  \put(62.2,0){\scriptsize \shortstack[c] {DANet\\ \cite{zhao2020DANet}}}
  \put(67.5,0){\scriptsize \shortstack[c] {PGAR\\ \cite{chen2020PGAR}}}
  \put(72.8,0){\scriptsize \shortstack[c] {$s^2$MA\\ \cite{liu2020S2MA}}}
  \put(78.2,0){\scriptsize \shortstack[c] {UC-Net\\ \cite{zhang2020ucnet}}}
  \put(83.5,0){\scriptsize \shortstack[c] {SSFRGBD\\ \cite{zhang2020select}}}
  \put(90,0){\scriptsize \shortstack[c] {JLDCF\\ \cite{Fu2020JLDCF}}}
  \put(95,0){\scriptsize \shortstack[c] {A2dele\\ \cite{piao2020a2dele}}}
  \end{overpic}
  \caption{Qualitative comparison against state-of-the-art RGB-D methods. (GT: ground truth)}
  \label{visualcmpRGBD}
  \vspace{-0.3cm}
\end{figure*}

We give more visual comparison results with the state-of-the-art RGB and RGB-D SOD methods in Figure~\ref{visualcmpRGB} and Figure~\ref{visualcmpRGBD}, respectively. It shows that our VST model can handle well in many challenging scenarios, \ie, big salient objects, cluttered backgrounds, foregrounds and backgrounds with very similar appearance, etc, while existing methods are heavily disturbed in these scenarios.
Besides, we also show the boundary maps predicted by our RGB VST and RGB-D VST models in Figure~\ref{visualcmpRGB} and Figure~\ref{visualcmpRGBD}, respectively. It can be seen that our models can predict clear boundaries for salient objects.

\end{document}